\documentclass[final,twocolumn]{elsarticle}

\usepackage{hyperref}
    \hypersetup{
        colorlinks   = true,
        citecolor    = blue
    }
\usepackage{lineno}
\usepackage{authorbiography}
\usepackage{cite}
\usepackage[utf8]{inputenc}
\usepackage[T1]{fontenc}
\usepackage{amsmath,amssymb,amsfonts}
\usepackage{algorithmic}
\usepackage{graphicx}
\usepackage{textcomp}
\usepackage{enumitem}
\usepackage[caption=false,font=footnotesize]{subfig}

\usepackage{color}
\usepackage{soul} 
 \usepackage{epsfig}

\graphicspath{{images/}}

\newcommand{\floor}[1]{\lfloor #1 \rfloor}

\newcommand{\DD}{\textbf{D}}
\newcommand{\II}{\textbf{I}}
\newcommand{\MM}{\textbf{M}}
\newcommand{\ASPADUPLA}[1]{``#1''} 
\newcommand{\PRT}[1]{\ensuremath{\left( #1 \right)}} 
\newcommand{\BRR}[1]{\ensuremath{\left| #1 \right|}} 

\journal{Neurocomputing}

\begin{document}
\begin{frontmatter}

\title{Dynamic Gesture Recognition  by Using CNNs and star RGB: a Temporal Information Condensation}

\author{Clebeson Canuto dos Santos}
\ead{clebeson.canuto@gmail.com}

\author{Jorge Leonid Aching Samatelo}
\ead{jorge.samatelo@ufes.br}

\author{Raquel Frizera Vassallo}
\ead{raquel@ele.ufes.br}

\address{Room 20, CT-II, Department of Electrical Engineering, Federal University Esp\'irito Santo, Av. Fernando Ferrari, 514, Goiabeiras, Vitória - ES - Brazil, ZIP code 29075-910}

\begin{abstract}
Due to the advance of technologies, machines are increasingly present in people's daily lives. Thus, there has been more and more effort to develop interfaces, such as dynamic gestures, that provide an intuitive way of interaction. Currently, the most common trend is to use multimodal data, as depth and skeleton information, to enable dynamic gesture recognition. However, using only color information would be more interesting, since RGB cameras are usually available in almost every public place, and could be used for gesture recognition without the need of installing other equipment. The main problem with such approach is the difficulty of representing spatio-temporal information using just color. With this in mind, we propose a technique capable of condensing a dynamic gesture, shown in a video, in just one RGB image. We call this technique star RGB. This image is then passed to a classifier formed by two Resnet CNNs, a soft-attention ensemble, and a fully connected layer, which indicates the class of the gesture present in the input video. Experiments were carried out using both Montalbano and GRIT datasets. For Montalbano dataset, the proposed approach achieved an accuracy of $94.58\%$. Such result reaches the state-of-the-art when considering this dataset and only color information. Regarding the GRIT dataset, our proposal achieves more than $98\%$ of accuracy, recall, precision, and F1-score, outperforming the reference approach by more than $6\%$.

\end{abstract}

\end{frontmatter}


\section{Introduction}

In Human-Machine Interaction (HMI), several communication interfaces can be used, such as manual controls, brain-reading devices, speech, gestures, and so on. In particular, gestures and speech are the least intrusive and most natural alternatives. Between speech and gestures, there are many cases where gestures are preferred rather than speech because they do not demand a mandatory grammar component for their elaboration \citep{mcneill1992hand}.

Gestures can be classified as static, i.e., there is no change of form over time, or dynamic, in which case there is a set of static gestures (poses) that vary within a time interval.
According to \citep{mcneill1992hand}, natural gestures, the most intuitive ones, are mostly dynamic and performed by hands. Therefore, when attempting to effectively use gesture for HMI, the static ones play a less significant role.
Due to the importance of gestures for HMI, several studies have been focused on the development of recognition systems that are more efficient. A list of the most used recognizers can be seen in \citep{mitra2007survey, rautaray2015vision, dynamic2016survey}, where \citep{dynamic2016survey} shows only those intended to recognize dynamic gestures.

Some the works mentioned above use more than one information in order to achieve better results, in which cases they are called multimodal methods~\citep{neverova2016moddrop}. They usually apply color (RGB format), depth measurements, and skeleton joints to detect and recognize gestures. This multiple-source information, such as depth and skeleton joints, complements positively the color information, which facilitates the separation of classes, to be done by the recognizer~\citep{multimodal2017challenges}.
However, to acquire depth information in  interactive environments, as well as other information besides RGB, it is usually necessary to have specific sensors, for example \textit{Microsoft Kinect}\footnote{\url{https://msdn.microsoft.com/en-us/library/hh438998.aspx}}, \textit{Asus Xtion Pro} \footnote{\url{https://www.asus.com/3D-Sensor/Xtion_PRO/}} or \textit{Intel Realsense} \footnote{\url{https://click.intel.com/realsense.html}}.

Such dependence on specific sensors causes a restriction on the interactional environment. Therefore, environments in which there are already RGB cameras, such as surveillance cameras that are easily found in many private or public spaces, can not be considered by such recognizers. This is one of the reasons that motivate studies aimed at the development of recognition methods based only on color information.

Some works use color as the only source of information, but just a few of them obtain significant results, such as \citep{barros2014real,multimodal2017challenges}. Even so, the good results mentioned in those works are achieved using particular gesture vocabularies, which contain gestures that are significantly different from each other, which usually simplifies the gesture recognition task. Thus, to improve HMI, we suppose that it is necessary to develop new techniques capable of recognizing dynamic gestures based only on color information, and that can deal with gestures that are not easily distinguishable from one another.

Recently, the use of Deep Learning (DL) has produced results situated in the state-of-the-arte of several problems within the areas of image processing and computer vision~\citep{liu2017survey,guo2016deep}. 
Among the architectures used in DL, Convolutional Neural Network (CNN) is currently the architecture achieving the best results in computer vision. This type of network has the property of sharing weights, which allows understanding the relationship between weights and input data as convolution operations. In this way, CNNs have several filters that can be trained to extract specific features from the data. 

As mentioned previously, recognizing dynamic gestures is not a simple task due to its temporal nature.
Therefore, even when using DL, sophisticated techniques are needed to capture temporal information. For instance, complex DL architectures as 3D-convolution~\citep{3dConvolution_for_HAR,3d_har2018}, Two-streams~\citep{zisserman2014, zisserman2016} or Two-streams with 3D-convolution~\citep{zisserman2017} are often used to perform action recognition.
These types of approaches require more processing power, implying a major constraint on their use in environments where real-time response is a priority.

Therefore, to tackle this problem, the authors of \citep{barros2014real} used a technique (called here \textit{star representation}) capable of condensing gestures temporal information in a single grayscale image. To this end, they used  a motion history image estimated by the sum of the absolute values of differences between consecutive frames. With this approach, the problem of gesture recognition in videos can be seen as an image classification problem. Hence, transfer learning can be used to recognize dynamic gestures by applying pretained CNNs, that is, CNNs that were previously trained for another image recognition task. Even though this approach can be interesting for gesture recognition, it has two  main drawbacks: 
(\textit{i}) gestures that can be distinguished only by its temporal sequence are particularly challenging to recognize, since the temporal information was reduced entirely to a spatial representation; 
(\textit{ii}) the process results in just one grayscale image, which may hamper the use of transfer learning, given that pretrained CNNs often receive as input an image with three channels as input (for example, an RGB image).

In order to solve these problems, we propose a dynamic gesture recognizer based exclusively on color information. Therefore, the contributions of this work are:
(\textit{i}) a new star representation for  gestures calculated from each input video, which  is an RGB image and can also encode more temporal information;
(\textit{ii}) a dynamic gesture classifier based on DL architecture composed by an ensemble of CNNs retrained and fused by a soft-attention mechanism. Even knowing that the result of an ensemble of models is usually better than an individual one\citep{goodfellow2016deep},  the contribution here is to show that soft-attention is a better option for fusing features from different CNNs than simple sum, mean or even concatenation.

Therefore, aiming to detail and explain our proposal, this paper is structured as follows: 
Section \ref{sec:relatedWorks} brings the related works and a brief explanation about the original star technique used to represent temporal information of gestures; 
Section \ref{sec:proposal} describes our  proposal; 
Section \ref{sec:experiments} presents our experiments; 
Section \ref{sec:discussion} discusses the results 
and finally, in Section \ref{sec:conclusions}, our conclusions and future works are exploited.

\section{Related Works}
\label{sec:relatedWorks}

Dynamic gesture recognition is a research field of increasing interest in recent years. Several previous works have focused on trying to develop a more intuitive interface to interact with machines and other devices. However, because a gesture can be represented in different ways within the same context, recognizing a dynamic gesture is a challenging issue. Besides, the way gestures are performed depends not only on the sequence of body movements but also on the cultural aspect of the people who make the gestures. Consequently, there is still much work to be done in order to obtain an interface capable of providing effective communication between humans and machines based on gestures.

For that reason, well-structured datasets that represent a variety of gesture meanings are of paramount importance for the development of this research area. 
Competitions like Chalearn: looking at people (LAP), cast a challenge for gesture recognition by releasing the Montalbano datasets V1~\citep{escalera2013multi} and V2~\citep{escalera2014}, that fulfill some of the aforementioned requirements. 
These datasets comprise approximately 14000 gestures shown in videos taken from 27 different subjects for 20 distinct classes.  
The sensor used to capture the data is a Microsoft Kinect 360. Thus, the data has multimodal nature since it consists of RGB, depth, user mask, 3D skeleton joints, and audio. The main difference between V1 and V2 versions, besides an improved annotation in the second version, is the lack of audio information, which is present only in the first one.

Even though Montalbano V1 and V2 are not datasets focused on HMI, due to their size and numerous samples, many works test their approaches using these two datasets, such as \citep{neverova2016moddrop,Efthimiou2016,LiChuankun2017,Joshi2017,Wang2017}. 
Analyzing such works, they can be separated into two main groups. First, the works that aim to solve the original problem of the challenge (recognizing a sequence of dynamic gesture from a video) and second, the works that classify the segmented gestures by using the labels available in the datasets. Both groups may recognize the gestures by using either multimodal data or only one sort of information.

Among the first group, which tried to solve the original challenge problem, most approaches have used multimodal information. For instance, \citep{neverova2016moddrop} has used all information available on both datasets. They applied all the different information to a set of CNNs. As a result, the authors obtained an accuracy of $96.81\%$ when they used all the data plus the audio, $93.1\%$ using only 3D skeleton and $95.06\%$ using RGB and Depth (RGB-D).

Some works use other different sources of data aiming at the same goal. For example, \citep{Neverova2015} use depth and skeleton joints to achieve a Levenshtein Distance (LD) of $0.85$, while \citep{Georgios2014} use skeleton joints, RGB and audio, reaching $0.88$. 

In \citep{Pigou2018}, the authors applied RGB and depth information as inputs to a CNN followed by an LSTM classifier~\citep{lstm}. With this architecture, they achieved a Jaccard Index (JI) of $0.906$ using RGB-D information, and $0.842$ using only RGB. 
One restriction with this approach is that the gesture should be made within a time window of $64$ frames. Consequently, when the gesture is performed in more or less than $64$ frames, the model likely does not work correctly. 

For the second group, usually the segmented gestures and their labels, provided with each dataset, were used to train and test the approaches in order to classify the gestures. Differently from the first group, their primary objective is to classify (recognize), without tackling both the problems of spotting and recognizing the gestures in a video sequence.

Among this group, the work presented by \citep{LiChuankun2017} achieved $91.6\%$ of accuracy using only the skeleton information. The authors represent the skeleton joints as a vector, and the sequence of skeletons as an image, where the $(x,y,z)$ coordinates of each joint represent a ($R$,$G$,$B$) pixel in the image. Using different representations of skeleton information, \citep{liu20193d,liu2019hidden} achieved  respectively $93.2\%$ and $93.8\%$. On the other hand, in \citep{ChenXi2014}, the authors used depth and skeleton information to classify the gestures, achieving only $85.5\%$ of accuracy. Other works such as \citep{ChenXi2014,Yao2014,Wu2014,Fernando2015,Escobedo-Cardenas2015,Joshi2017} also used different types of information, but did not achieve a better result than \citep{liu2019hidden}. Besides, in these works, the videos were cut into clips containing only the gestures. This clipping process was implemented using the labels related to each gesture present in the dataset. It is important to mention that skeleton joints are a data source that condenses dynamic and structure information related to the gesture. Consequently, it is suitable to be used for dynamic gesture recognition. However, this type of data is usually provided by specific sensors, like the \textit{Microsoft Kinect}, which was adopted to acquire the mentioned dataset. Table \ref{tab:montalbano-works} shows a summary of the best results achieved over Montalbano datasets for different types of data.

Unfortunately, Kinect-like sensors are not easily found in ordinary places, which usually have regular surveillance cameras. Therefore, approaches that use only RGB images are most preferred than multimodal methods, that need more than one type of information in order to work.

 \begin{table}[!h]
 \caption{The main works that use the Montalbano datasets}
 \label{tab:montalbano-works}
 \centering
\begin{tabular}{lllll}
\hline
\multicolumn{1}{c}{\textbf{Work}} & \multicolumn{1}{c}{\textbf{Used Data Type}} & \multicolumn{1}{c}{\textbf{Result}} \\
\hline
\citep{neverova2016moddrop}   & Audio, RGB-D and Skeleton    &    96.81\%\\ 
\citep{neverova2016moddrop}   & RGB-D   &    95.06\%\\ 
\citep{neverova2016moddrop}   & Audio    &    94.96\%\\ 
\citep{liu2019hidden} & Skeleton & 93.8\%\\
\citep{Efthimiou2016}   & Audio and RGB    &  93.00\%\\
\citep{Escobedo-Cardenas2015}   &  Skeleton and RGB-D &  88.38\%\\
\citep{Wu2016}    & Depth    & 82.62\%\\
\citep{Fernando2015}   & Skeleton and Audio &  80.29\%\\
\citep{CongqiCao2015}   & RGB  & 60.07\%\\

\hline
\end{tabular}
\end{table}

Nevertheless, until the elaboration of this paper, works that perform gesture recognition using only RGB information reach at most $60\%$ of accuracy in the Montalbano datasets. Two possibilities can explain these results: either the approaches proposed were focused on recognizing gestures using only multimodal information and did not make a significant effort on using just one type of information, or the gestures are too difficult to distinguish one from another when using exclusively the color information.

Considering only RGB images, the authors from \citep{barros2014real} used a variant of the Motion History Image (MHI) technique~\citep{MHIrepresentation} to represent the movement information contained in a video sequence. This representation, denoted here as star, was used alongside with a CNN to recognize dynamic gestures. The results reported using a dataset developed by authors was of $91.67\%$ of accuracy. It is relevant to mention that the gestures of this dataset are very different from each other and therefore, the classes can be easily recognized.
Nevertheless, we consider that the star representation is particularly interesting and can be applied in more challenging datasets. As such, in the next paragraphs, we will give a detailed description of the star representation calculation.

Considering a grayscale video with $N$ frames containing a dynamic gesture, the star representation can be calculated with two main steps, as follows:

\noindent \textbf{First step}, the accumulated sum of the absolute difference between consecutive video frames is calculated. The Equation \eqref{eq:star1} represents this process.

\begin{equation}
\label{eq:star1}
\MM(i,j) = \sum_{k=2}^{N}|\II_{k-1}(i,j) - \II_{k}(i,j)|W_s,
\end{equation}

\noindent where: 
$N$, called memory size, is the number of frames of each video clip containing a gesture; 
$(i,j)$ are the coordinates of a pixel in a frame; 
$\MM$ is the star image; 
$\II_{k}$ is the $k^{th}$ frame of a video containing a dynamic gesture; 
$|\bullet|$ is the modulus operator and 
$W_s=k/N$ is called weighted shadow, and it is responsible for weighting the absolute difference depending on $N$.

\noindent \textbf{Second step}, Sobel masks are applied over the matrix $\MM$ in the $X$ and $Y$ directions, obtaining the matrices $\MM_X$ and $\MM_Y$, respectively. Finally, the star representation is defined as three grayscaled images, corresponding to the matrices $\MM$, $\MM_X$ and $\MM_Y$. 

According to the authors, the two extra channels $\MM_X$ and $\MM_Y$ provide better discrimination about the type of movement present in $\MM$. However, when using a CNN network, usually the first convulational layers learn and assume the function of these types of masks, during the training stage. That eliminates the need of applying the Sobel masks over the matrix $\MM$.

With this in mind, this work proposes some changes over the star representation in order to generate an RGB image that encodes the gesture movement, contained in a video. The new representation is applied to a DL architecture based on an ensemble of two CNNs fused by a soft-attention mechanism.  This proposal is evaluated over the Montalbano and GRIT~\citep{tsironi2017analysis} datasets, achieving higher accuracy in the gesture recognition process in both of them. As such, the two main contributions of this work are the changes  proposed in the star representation; as well as the DL architecture used to solve the gesture recognition problem.

\section{Proposal}
\label{sec:proposal}

In this section, the proposal for our dynamic gesture recognition technique is described. Two main steps form our proposal:

\begin{itemize}
\item
\textbf{Pre-processing}: each input video is represented as an RGB image by using a modified version of the aforementioned star representation. 
\item
\textbf{Classification}: a dynamic gesture classifier is trained using an ensemble of CNNs. Specifically, the image obtained from the pre-processing step is passed as input to two pre-trained CNNs. The results of these two CNNs pass through a soft-attention mechanism, and then, after being weighted, the results are sent to a fully connected layer. Finally, a softmax classifier indicates the class to which the gesture may belong.
\end{itemize}

Details of these two steps are given in the following subsections.

\subsection{Pre-processing}
\label{entry}

The original star representation does not take into account the color information of each frame. Consequently, the resulting representation is a grayscale image, calculated using Equation \eqref{eq:star1}. Thus, our first goal is to represent temporal information present in a color video by using an improved version of the proposal in~\citep{barros2014real}.

Therefore, to take advantage of the color information, the difference between two consecutive frames, calculated as  Equation~\eqref{eq:star1} in \citep{barros2014real}, can be replaced by the Euclidean Distance as presented in Equation~\eqref{eq:euclidian}.

\begin{equation}
\MM(i,j) = \sum_{k=2}^{N}||\II_{k-1}(i,j) - \II_{k}(i,j)||_2,
\label{eq:euclidian}
\end{equation}

\noindent where $\II_{k}(i,j)$ represents the RGB vector of a pixel at the position $(i,j)$ of the $k^{th}$ frame; and $||\bullet||_2$ represents the $L_2$ norm.

However, the Euclidean Distance considers only the vector norm and can only evaluate the intensity of each image. Thus, a better solution would be to include information from both magnitude and phase when calculating the distance between RGB vectors. That would allow us to  evaluate not just the changes in image intensity, but the hue and saturation as well. In this sense,  \citep{samatelo2012new} proposed a metric based on cosine similarity, described as shown in  \eqref{eqANALISISINI002} and \eqref{eqANALISISINI006B}. Such metrics is the one used in our work, in order to build a modified version of the star representation.

\begin{equation}
\lambda = 1 - \cos(\theta) = 1 - \frac{\II_{k-1}(i,j)^T\II_{k}(i,j)}{||\II_{k-1}(i,j)||_2||\II_{k}(i,j)||_2},
\label{eqANALISISINI002}
\end{equation}

\noindent where $\theta$ is the angle between  $\II_{k-1}(i,j)$ and $\II_{k}(i,j)$. 

\begin{equation}
\label{eqANALISISINI006B}
\DD_k(i,j) = \PRT{1 - \frac{\lambda}{2}}.\BRR{||\II_{k-1}(i,j)||_2 - ||\II_{k}(i,j)||_2}.
\end{equation}

Since $\DD_k(i,j) \approx ||\II_{k-1}(i,j) - \II_{k}(i,j)||_2$,  we can use Equation \eqref{eqANALISISINI006B} in Equation  \eqref{eq:euclidian}, becoming, finally, Equation \eqref{eq:star2}) , which is the proposed star representation. This new equation substitute the original star representation, given by Equation \eqref{eq:star1}.

\begin{equation}
\label{eq:star2}
\MM(i,j) = \sum_{k=2}^{N}\DD_k(i,j).
\end{equation}

\begin{figure}[!h]
  \centering
  \subfloat[]{\includegraphics[width=0.3\linewidth]{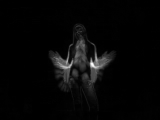}}
  \subfloat[]{\includegraphics[width=0.3\linewidth]{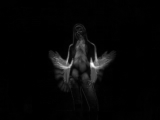}}
  \subfloat[]{\includegraphics[width=0.3\linewidth]{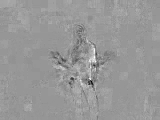}}
  
  \caption{Comparing the approaches. a) star representation proposed by~\citep{barros2014real} b) our modified star representation c) the difference between them.}
\label{fig:star_diff}
  \end{figure}

From Equation~\eqref{eq:star2}, the distance between two consecutive images can be calculated using the difference of intensities, scaled by a numeric value that depends on the angle between each RGB pixel of the images. Therefore, both intensity and chromaticity information are taken into account. Figure~\ref{fig:star_diff} shows the results of Equation \eqref{eq:star1} (star representation) and Equation  \eqref{eq:star2} (modified star representation).

It is possible to notice that with the new proposal, more information of the movement can be extracted. However, even with this new representation, the result is still a grayscale image and, which does not match with state-of-the-art CNN architectures, that usually receives an RGB image as input. Besides, this approach still does not solve the problem of temporal information loss. Therefore, movements with the same path but performed with different directions will have similar representations.

\begin{figure} [h!]
\centering

\includegraphics[width =1.0\linewidth]{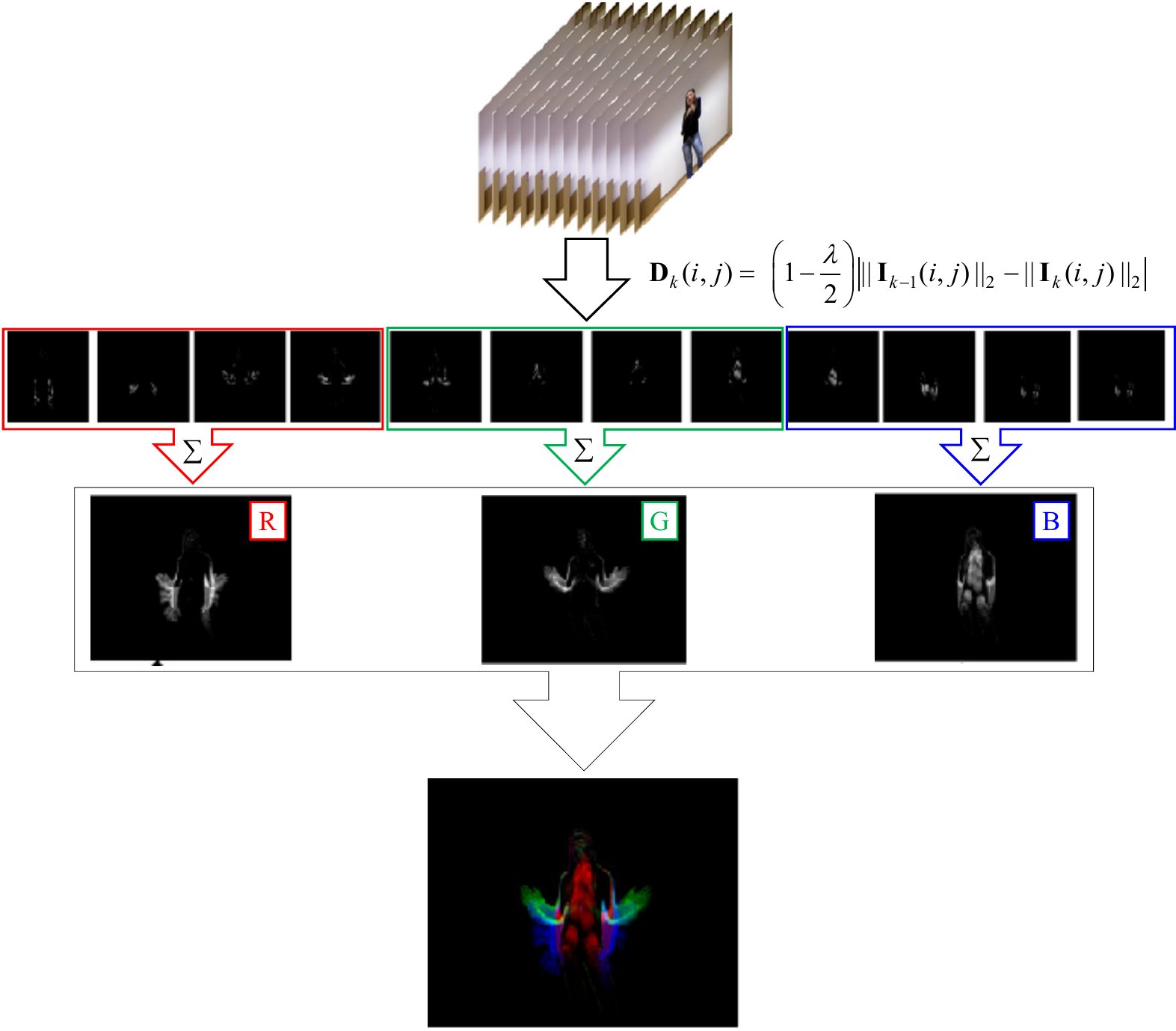}
\caption{star RGB representation for the gesture \textit{basta}.}
\label{fig:star_rgb}
\end{figure}

Thus, to improve the temporal representation and, simultaneously, create an RGB image as an output, we propose an approach summarized in Figure~\ref{fig:star_rgb} and explained below:

\begin{itemize}
\item

Each color video containing a complete dynamic gesture is equally divided into three sub-videos of $\floor{\frac{N}{3}}$ frames, which are assumed to be the pre-stroke, stroke and post-stroke steps of a dynamic gesture, as defined in~\citep{mcneill1992hand} and discussed by~\citep{liu2019hidden}. If the number of frames is not divisible by three, the middle sub-video will contain $N - 2\floor{\frac{N}{3}}$ frames;
\item
For each resulting sub-video, the matrix $\MM$ of the star representation is calculated using Equation~\eqref{eq:star2}.
Then, from the three $\MM$ matrices, the entire video is represented by an RGB image, where the $R$-channel contains the $\MM$ matrix calculated from the first sub-video, the $G$-channel has the $\MM$ matrix from the central one, and the $B$-channel the $\MM$ matrix from the last one.
\end{itemize}

Because the output of our approach is an RGB image, such image is called star RGB representation.

\begin{figure}[!h]
  \centering
  \subfloat[]{\includegraphics[width=0.3\linewidth]{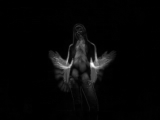}\label{fig:star_diff_oposite_A}}
  \subfloat[]{\includegraphics[width=0.3\linewidth]{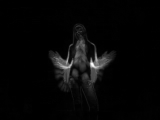}\label{fig:star_diff_oposite_B}}
  \subfloat[]{\includegraphics[width=0.3\linewidth]{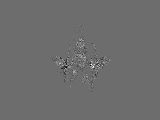}\label{fig:star_diff_oposite_C}}
  
  \subfloat[]{\includegraphics[width=0.3\linewidth]{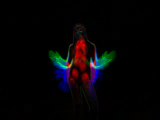}\label{fig:star_diff_oposite_D}}
  \subfloat[]{\includegraphics[width=0.3\linewidth]{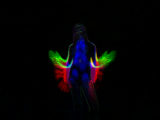}\label{fig:star_diff_oposite_E}}
  \subfloat[]{\includegraphics[width=0.3\linewidth]{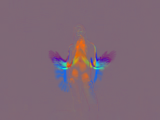}\label{fig:star_diff_oposite_F}}
  
\caption{Comparing results between star and star RGB approaches. 
a) star representation of a video in its original frame sequence; 
b) star representation of a video in its inverted frame sequence; 
c) the difference between a) and b); 
d) star RGB of a the video in its original frame sequence; 
e) star RGB of a the video in its inverted frame sequence; 
f) the difference between d) and e). Note that image c) is almost zero. Most of the pixel values are close to zero (something about 1e-7) and were normalized so we could see the difference between images (a) and (b).}
\label{fig:star_diff_oposite}
\end{figure}

Besides resulting in a multichannel and sparse representation of a color video, we observe that the star RGB representation has another advantage when the direction of the movement is required for the gesture to be distinguished, like waving to someone to move in a specific direction. This consideration is illustrated by simulating two gestures with the same movements but with opposite directions. For instance, Figure~\ref{fig:star_diff_oposite} shows the star and star RGB representations calculated from two videos, where the second video is generated by reversing the order of the frames of the first one. Hence, if we compare Figures~\ref{fig:star_diff_oposite_A} and ~\ref{fig:star_diff_oposite_B}, we can notice that the original star representation produces similar grayscale images for both videos, which is evident from Figure~\ref{fig:star_diff_oposite_C} that shows the difference between them. In contrast, the star RGB produces two different color images for each video, as shown in Figures ~\ref{fig:star_diff_oposite_D} and ~\ref{fig:star_diff_oposite_E}, while Figure~\ref{fig:star_diff_oposite_F} shows the difference between each channel of the images~\ref{fig:star_diff_oposite_D} and \ref{fig:star_diff_oposite_E} in RGB format.

This result suggests that the star RGB improves the original representation, encoding more temporal information than the previous approach presented in~\citep{barros2014real}. Besides, it probably makes the classifier model equivariant to similar movements that represent different gestures.

In summary, star RGB represents the temporal information of a video better by associating the color channels with the pre-stroke, stroke and post-stroke moments of a dynamic gesture. In addition, it is more suitable for training models based on state-of-the-art CNNs aimed at image classification. And, its calculation is easy and fast.

\subsection{Classification}
\label{sec:classification}

\begin{figure} [!h]
\centering
\includegraphics [width = \linewidth]{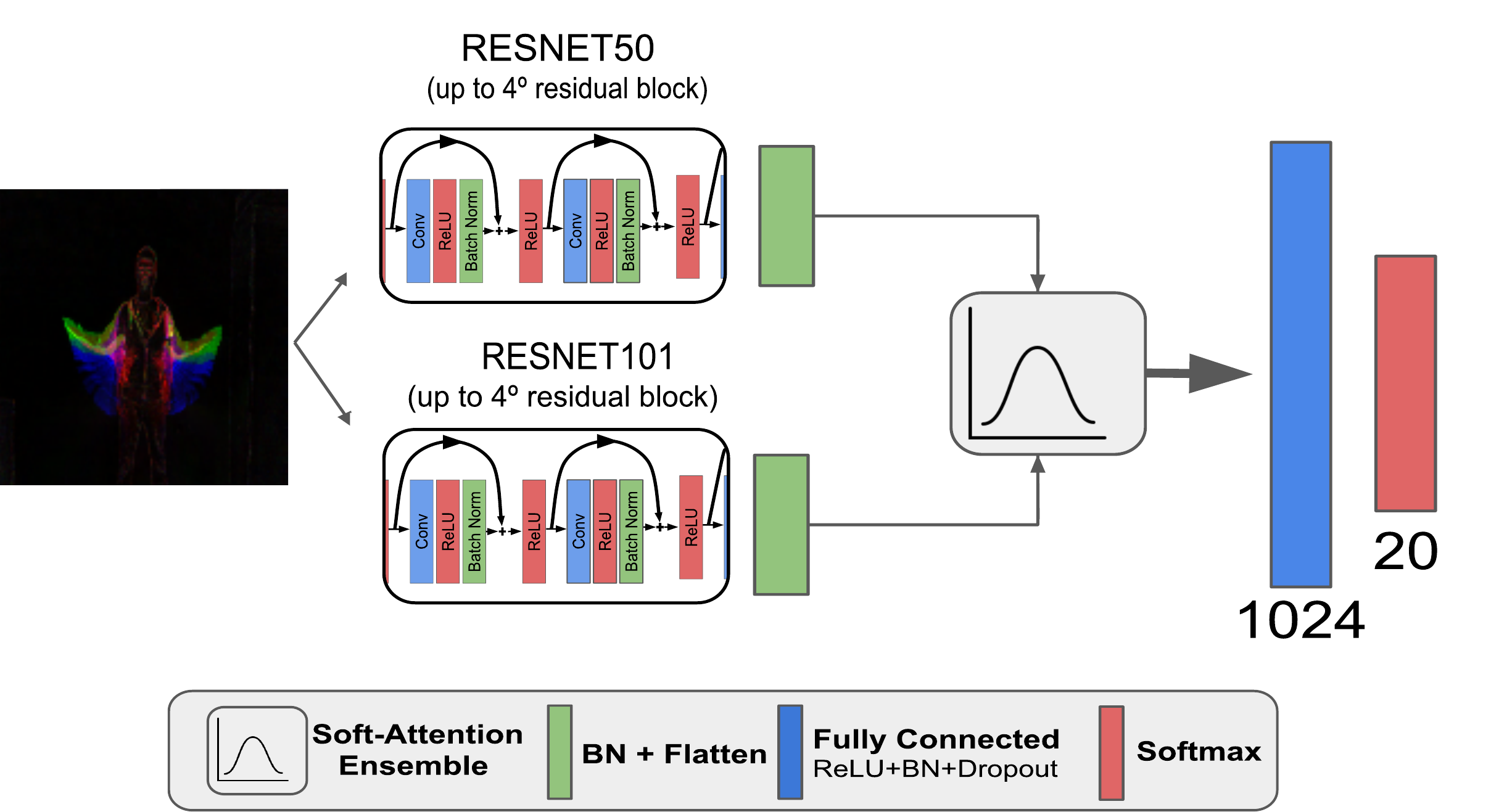}
\caption{Proposal of dynamic gesture classifier.}
\label{fig:ensemble}
\end{figure}

After pre-processing a video sequence to build the corresponding star RGB representation, the next step is classification. Our approach for the dynamic gesture classifier, shown in Figure \ref{fig:ensemble}, comprises three parts:
(\textit{i}) a feature extractor, based on pre-trained CNNs;
(\textit{ii}) an ensemble of CNNs, where features are fused by a soft-attention mechanism;
(\textit{iii}) and a classifier, formed by 2 fully connected layers, with the last one normalized by a softmax function. 
Each one of these parts is explained as follows.

\subsubsection{Feature extractor}

The feature extractor is based on the CNN Resnet~\citep{resnet}, which is specialized in image classification, and is previously trained using the dataset \textit{ImageNet}. Such dataset is released in the ILSVRC-2014 competition and contains more than $1.2$ million images distributed over $1000$ distinct categories~\citep{russakovsky2015imagenet}. Resnet is one of the most used CNNs to classify images. Moreover, this architecture achieves the state-of-the-art in several image classification problems. These excellent results are a consequence of the residual blocks, which can mitigate the vanishing gradient problem even in a very deep architecture~\citep{resnet}.

The Resnet is chosen  after an empirical selection procedure, where Resnet 101 and Resnet 50 obtained the best results. Such procedure evaluates the performance of seven distinct CNN architectures (Resnet (18, 34, 50, 101, 121), VGG16\citep{simonyan2014vgg} and DenseNet\citep{guo2016densenet}) when inputting the RGB image generated in the pre-processing step.

\subsubsection{Ensemble}

The feature maps corresponding to the $4^{th}$ convolutional layer of each Resnet are passed sequentially through an attention mechanism, referred here as the soft-attention ensemble. This mechanism plays a role of weighting each feature map according to its importance to the predicted class.

The soft-attention ensemble is chosen regarding the following considerations:
(\textit{i}) it should evaluate the information according to its importance for the task, differently from other fusion types as summation and arithmetic mean or even max-pooling;
(\textit{ii}) the sequential nature of the soft-attention ensemble avoids the concatenation of all feature maps of the inputs, which would increase the size of the input vector generated by the ensemble.

The architecture of the soft-attention ensemble is shared by all feature maps and is composed of a fully connected layer with 128 neurons using ReLU~\citep{relu2010rectified} as activation function, and an output layer of one neuron without an activation function. Thus, it receives a vector (a flattened CNN feature map) as input while outputting just one value.

\begin{figure}[!h]
\centering
\includegraphics [width = \linewidth]{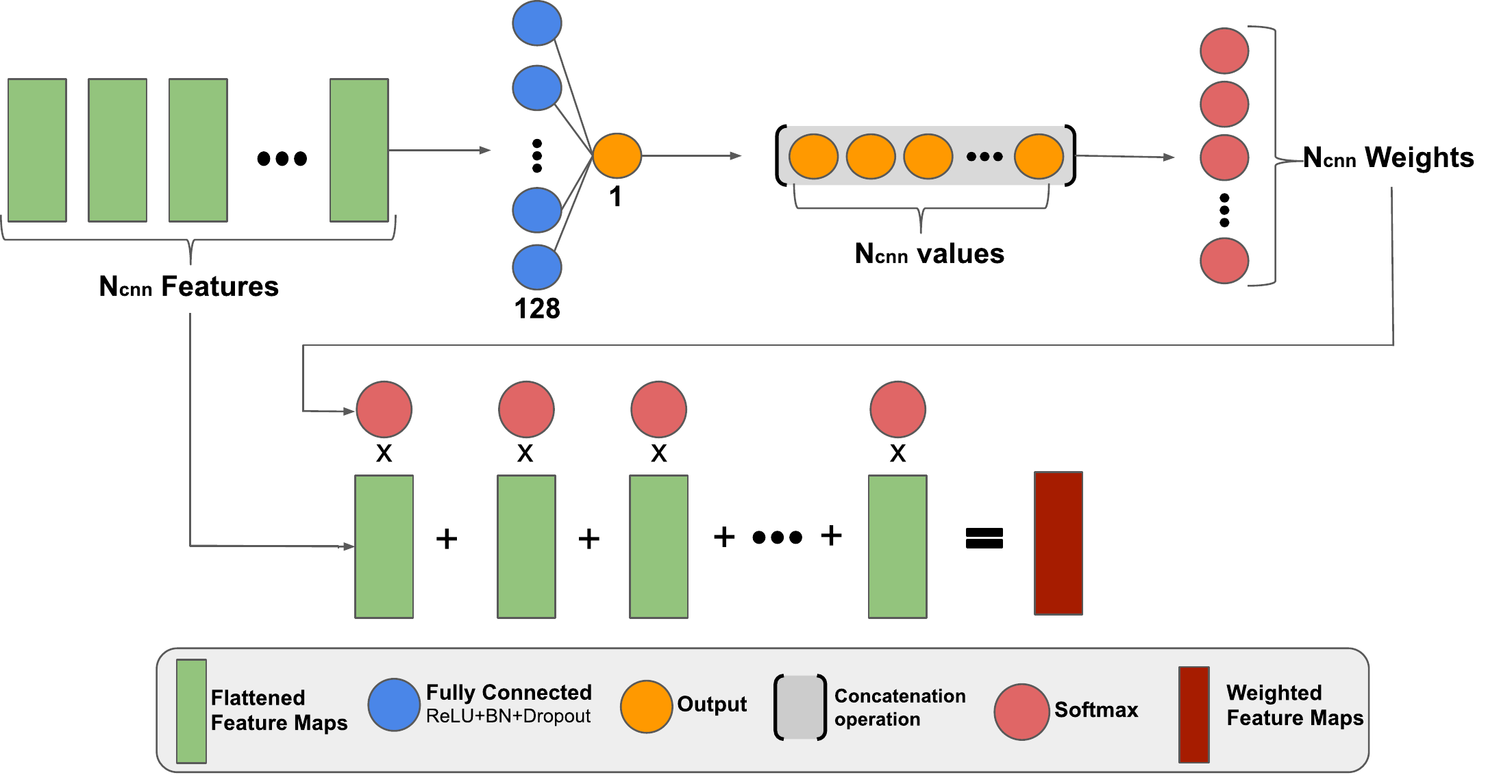}
\caption{Soft-attention ensemble.}
\label{fig:soft_ensemble}
\end{figure}

The sequential operation of the soft-attention ensemble is shown in Figure~\ref{fig:soft_ensemble}. Let $N_{CNN}$ be the number of CNN feature maps that will be fused by the soft-attention.
Firstly, each flattened CNN feature map is applied over the soft-attention step, generating a vector with $N_{CNN}$ elements, that is normalized using a softmax function.  
Second, the weighted sum of the $N_{CNN}$ flattened feature maps is calculated using the elements of the normalized vector as weighting coefficients. 
In this work, we use $N_{CNN} = 2$, given that the feature extractor is based on two CNNs. It is worth noticing that, when using this mechanism, all feature maps must have the same dimensions.

It is important to mention that Batch Normalization~\citep{ioffe2015batch} is applied before the feature maps are passed to the soft-attention step, which is required because the feature maps do not necessarily have the same distributions. Therefore, using the soft-attention mechanism directly would not give an accurate result when regarding the importance of each feature map for the problem.
 

\subsubsection{Classifier}

The output of the soft-attention ensemble feeds a classifier, which is composed of a hidden layer of 1024 neurons with Batch Normalization, dropout~\citep{srivastava2014dropout} and ReLU activation function; 
and an output layer of 20 neurons with a softmax activation function. The softmax activation function over the last layer gives the probability that the input image belongs to one of the $20$ gestures of the dataset.

In addition, the cost function used in this work is the mean of the Cross-entropy, calculated for each minibatch and regularized by the $L1$ norm of the entire set of weights, which includes those from the feature extractor, ensemble, and classifier. This type of regularization was applied to force the sparsity of the weights, what, in theory, can better handle sparse input images (see Figure~\ref{fig:star_rgb}).

\section{Experiments and results}
\label{sec:experiments}

This section discusses the datasets used throughout the experiments, the implementation and training of the proposed architecture and, finally, the results obtained in the evaluation step.

In short, two different experiments are performed. The first experiment is carried out to evaluate the proposed architecture and the impact of the soft-attention ensemble on the gesture classifier performance. The second experiment is aimed to evaluate the use of the star RGB representation within the proposed architecture. These experiments are conducted with different datasets, which is explained as follows.

\subsection{Montalbano gesture dataset}

\begin{figure} [!h]
\centering
\includegraphics [width = \linewidth]{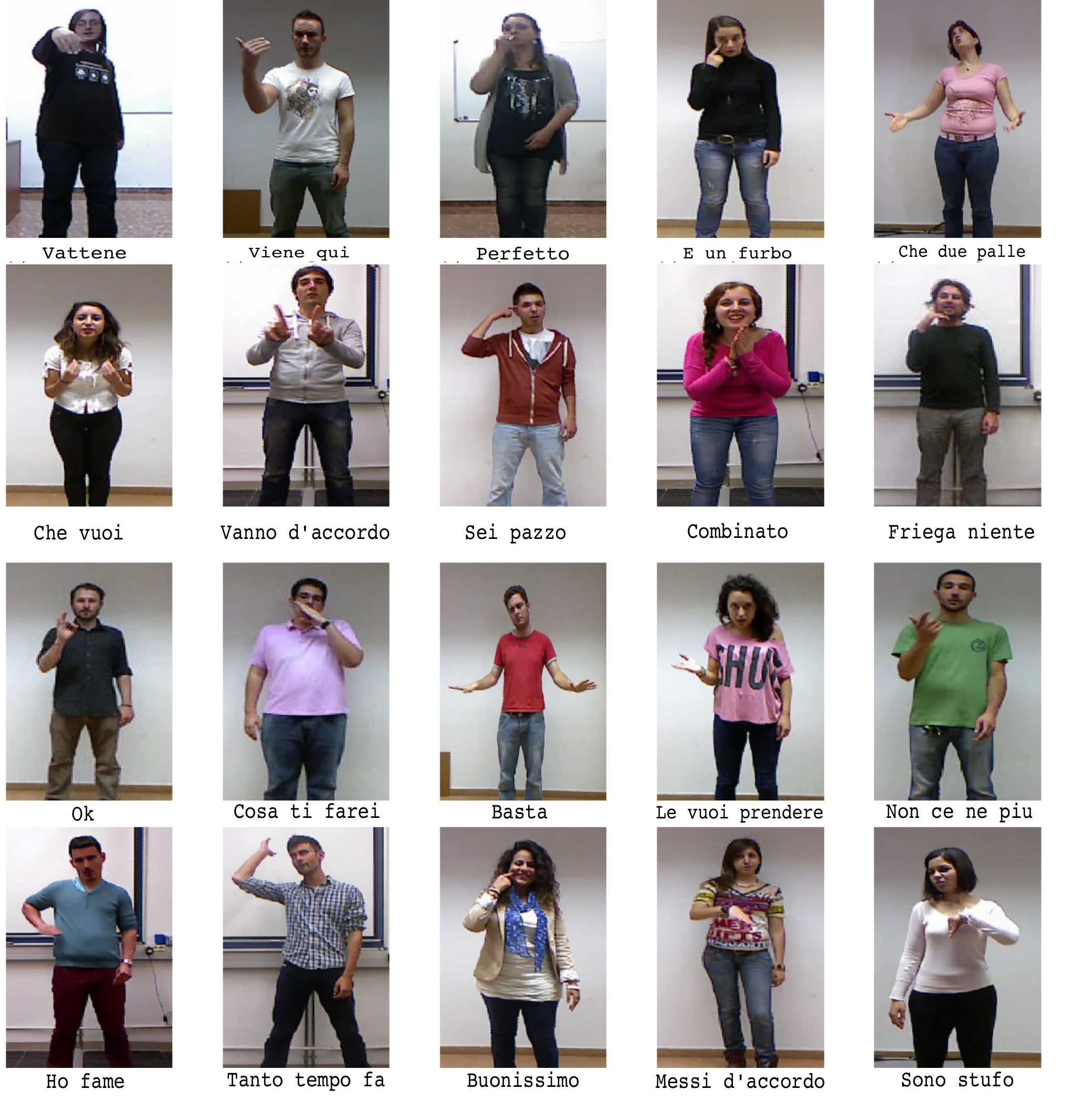}
\caption{ Representation of the 20 gestures present in the 
Montalbano gesture dataset~\citep{escalera2013multi}. The text under each figure represents the respective gesture name in Italian Language.}
\label{fig:database}
\end{figure}

This dataset was cast in the \textit{Chalearn: looking at people} - 2014 challenge~\citep{escalera2014}, and  comprises $13206$ cultural/anthropological Italian gestures ($6862$ for training, $2765$ for validation and $3579$ for testing), distributed over $20$ different classes. All gestures were captured using a Kinect $360$ sensor and present multimodal information: RGB, depth, skeleton and user mask. In the challenge, the candidates could use any data provided to recognize a sequence of gestures (between $8$ and $20$ per video) performed in a continuous video.
Figure~\ref{fig:database} shows a representation of the gestures contained in this dataset. 

This dataset is one of the largest set currently released that is focused on the problem of dynamic gesture recognition, which justifies its choice in our work.  Also, as mentioned by \citep{tsironi2017analysis}, the available datasets of dynamic gestures based on RGB are very small and are captured usually focusing only on the information of the hands rather than the entire body. 

Although the original challenge in~\citep{escalera2014} is based on using multimodal data to recognize dynamic gestures, we believe that using only RGB information is enough to sucessfully solve this problem.

Our main motivation comes from the fact that surveillance cameras can be easily found in public places, where gestures could be used to promote the interaction between humans, devices and, also, the environment. Therefore, being able to recognize gestures with the need of only color information is an attractive possibility for an HMI that may be deployed in places where standard cameras are installed.

Since the Chalearn Competition is not part of our goal, we separate the gestures from several videos, according to the provided labels. Thus, each video of training or test set contains only one dynamic gesture. That way, the problem is no longer to recognize a sequence of gestures in a video but to classify videos in $20$ different classes of dynamic gestures.

\begin{figure}[!h]
\centering
\subfloat[vattene]{\includegraphics[width=0.23\linewidth]{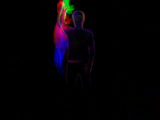}}\,
\subfloat[vieniqui]{\includegraphics[width=0.23\linewidth]{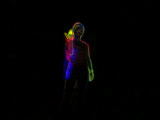}}\,
\subfloat[perfetto]{\includegraphics[width=0.23\linewidth]{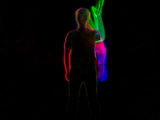}}\,
\subfloat[furbo]{\includegraphics[width=0.23\linewidth]{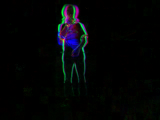}} \,

\subfloat[cheduepalle]{\includegraphics[width=0.23\linewidth]{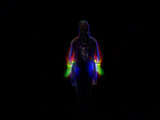}}\,
\subfloat[chevuoi]{\includegraphics[width=0.23\linewidth]{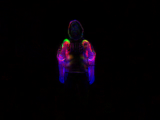}}\,
\subfloat[daccordo]{\includegraphics[width=0.23\linewidth]{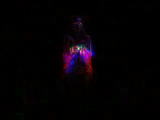}}\,
\subfloat[seipazzo]{\includegraphics[width=0.23\linewidth]{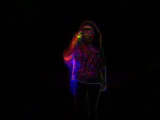}}\,

\subfloat[combinato]{\includegraphics[width=0.23\linewidth]{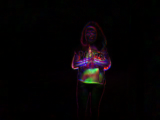}}\,
\subfloat[freganiente]{\includegraphics[width=0.23\linewidth]{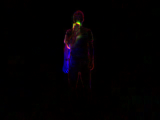}}\,
\subfloat[ok]{\includegraphics[width=0.23\linewidth]{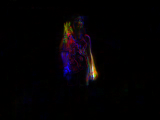}}\,
\subfloat[cosatifarei]{\includegraphics[width=0.23\linewidth]{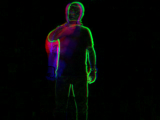}} \,

\subfloat[basta]{\includegraphics[width=0.23\linewidth]{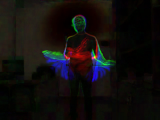}}\,
\subfloat[prendere]{\includegraphics[width=0.23\linewidth]{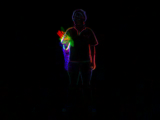}}\,
\subfloat[noncenepiu]{\includegraphics[width=0.23\linewidth]{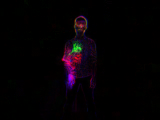}} \,
\subfloat[fame]{\includegraphics[width=0.23\linewidth]{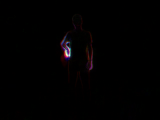}}\,

\subfloat[tantotempo]{\includegraphics[width=0.23\linewidth]{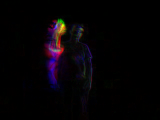}}\,
\subfloat[buonissimo]{\includegraphics[width=0.23\linewidth]{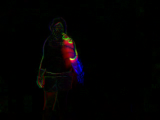}}\,
\subfloat[messidaccordo]{\includegraphics[width=0.23\linewidth]{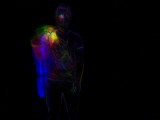}}\,
\subfloat[sonostufo]{\includegraphics[width=0.23\linewidth]{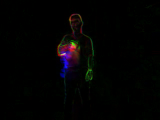}}\,

\caption{A star RGB representation of one sample of each gesture present in the Montalbano dataset.}
\label{fig:montalbano_star}

\end{figure}

After all the videos are segmented, the proposed technique is applied, as presented in (Section \ref{sec:classification}).  Figure~\ref{fig:montalbano_star} illustrates a sample of each gesture shown in Figure~\ref{fig:database} after calculating the star RGB representation.
 
\subsection{GRIT gesture Dataset}

\begin{figure}[!h]
\centering
\includegraphics[ width = \linewidth]{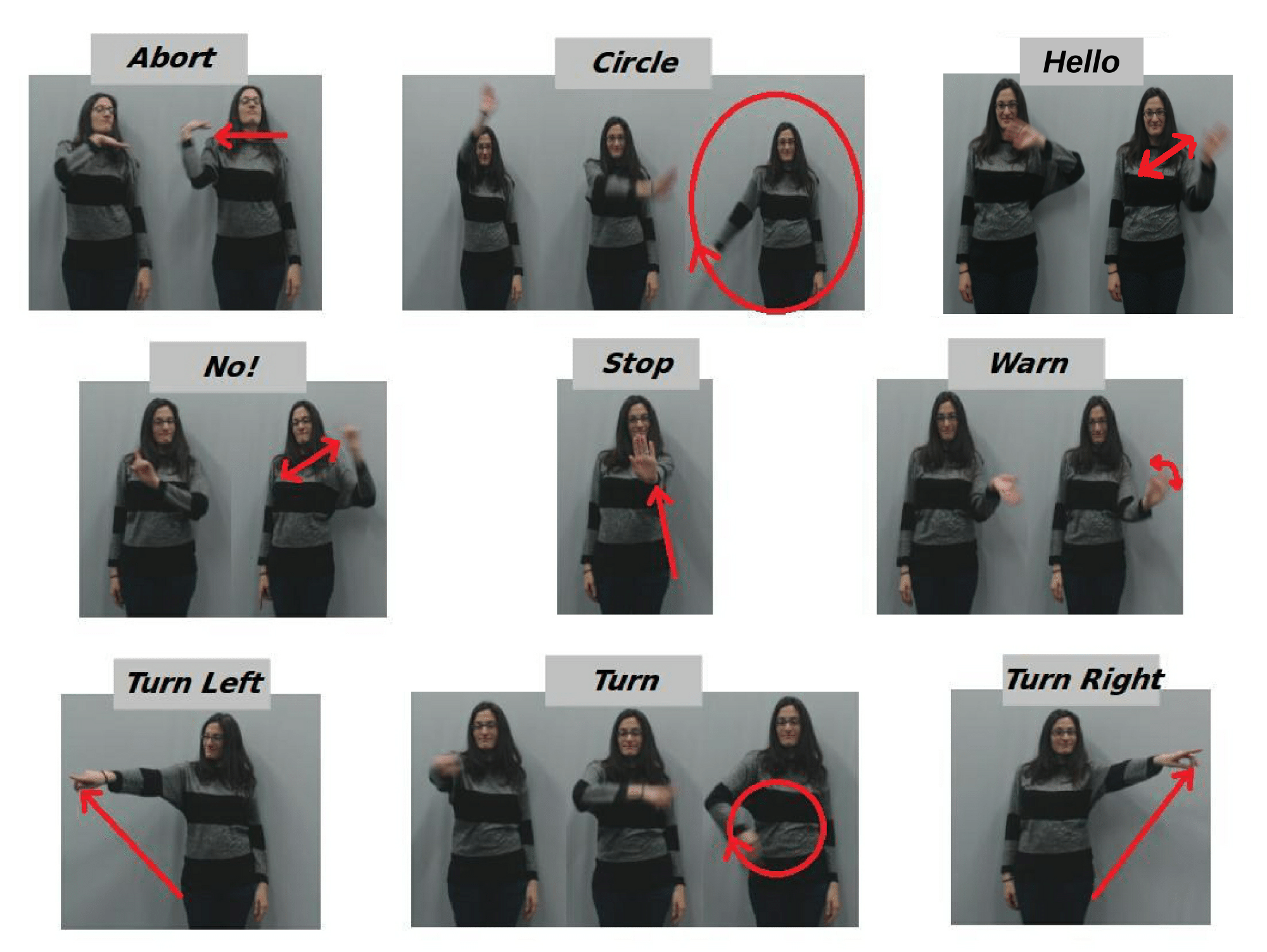}
\caption{Representation of the nine gestures in the  GRIT dataset~\citep{tsironi2017analysis}.}
\label{fig:grit_dataset}
\end{figure}

\begin{figure}[!h]
\centering
\subfloat[Abort]{\includegraphics[width=0.3\linewidth]{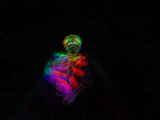}} \,
\subfloat[Circle]{\includegraphics[width=0.3\linewidth]{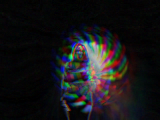}} \,
\subfloat[Hello]{\includegraphics[width=0.3\linewidth]{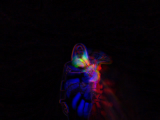}} \,

\subfloat[No!]{\includegraphics[width=0.3\linewidth]{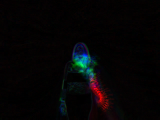}} \,
\subfloat[Stop]{\includegraphics[width=0.3\linewidth]{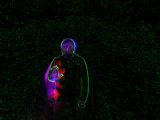}} \,
\subfloat[Warn]{\includegraphics[width=0.3\linewidth]{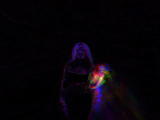}} \,
        
\subfloat[Turn left]{\includegraphics[width=0.3\linewidth]{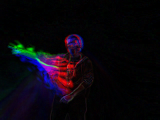}}\, 
\subfloat[Turn]{\includegraphics[width=0.3\linewidth]{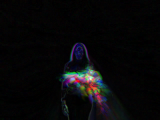}} \,
\subfloat[Turn Right]{\includegraphics[width=0.3\linewidth]{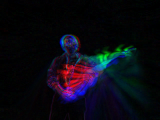}} 
        
\caption{The star RGB representation of one sample of each gesture of the GRIT dataset~\citep{tsironi2017analysis}.}
\label{fig:grit_star}

\end{figure}

In order to evaluate our proposal for recognizing gestures used in human-robot interaction, we consider the GRIT dataset (Gesture Commands for Robot InTeracton), which comprises $543$ gestures, distributed over nine distinct classes. Unlike the Montalbano dataset, this one contains one dynamic gesture per video.  Furthermore, as it was created to be used for interaction with robots, its gestures are quite separable. Figure~\ref{fig:grit_dataset} illustrates a representation of nine gestures available in the dataset. Additionally, Figure~\ref{fig:grit_star} illustrates the star RGB representation of a sample of each one of its nine gestures.

\subsection{Implementation and Training}

The proposed architecture is implemented using \textit{pytorch} V1.0, an open source software developed by \textit{Facebook's artificial-intelligence research group} \footnote{\url{https://research.fb.com/category/facebook-ai-research/}} for machine learning~\citep{pytorch}.

The computer used in the experiments has the following configuration:
(\textit{i}) Operating System Linux Ubuntu Server, distribution 16.04;
(\textit{ii}) Intel Core i7-7700 processor, 3.60GHz with 4 physical cores;
(\textit{iii}) $32$GB of RAM;
(\textit{iv}) 1TB of storage unit (hard drive);
(\textit{v}) Nvidia Titan V graphic card, with $12GB$ of dedicated memory.

During the training step, some techniques of data augmentation are applied. Specifically, for the Montalbano dataset, a random crop of size $110 \times 140$ pixels, random horizontal flip, random rotation with $\pm5^o$, and random Gaussian noise, with $\mu = 0$ and $\sigma = 1$, were applied. On the other hand, for the GRIT dataset, only a random crop was used.  In addition, Table \ref{tab:hyper} presents the hyper-parameters used during the training steps for both datasets.

\begin{table*}[!h] 
\centering
\caption{Hyper-parameters used during the training step by each dataset.}
\begin{tabular}{lcc}
\hline 
Hyper-parameters &  Montalbano & GRIT\\
\hline 
Batch size &  64  &    8     \\
Epochs &  100   &   100      \\
Learning Rate &  $1\mathrm{e}{-4}$ (CNN) &  $1\mathrm{e}{-4}$ (CNN) \\
& $1\mathrm{e}{- 3}$ (FC)   &  $1\mathrm{e}{- 3}$ (FC) \\
Learning Rate decay & 1\% per epoch    &  1\% per epoch       \\
Dropout (keep prob.) &  0.2  &    0.2     \\
Optimizer algorithm &   Adam  &  Adam       \\
\hline 
\end{tabular} 
\label{tab:hyper}
 \end{table*}

Finally, regarding the training step, an early stop strategy is adopted. Therefore, training is executed until an average accuracy greater than $99\%$ is achieved over the last $5$ epochs.

Due to implementation compatibility, it is necessary to resize each frame. Hence, regarding the training step, each frame is resized to $120 \times 160$ pixels before data augmentation, and for the test step, each frame is cropped at the center point, resulting in an $110 \times 140$ image.

\subsection{Performance evaluation}

For multiclass classification problems, it is common to analyze the result using accuracy, as metrics. In the context of gesture recognition, the accuracy rate is calculated as being the number of gestures correctly classified divided by the total number of gestures present in the test dataset. 

For the Montalbano dataset, the accuracy achieved by each class is  presented in tables. Futhermore, to improve the visualization of the behavior of the classifier, the results is also described by means of a confusion matrix, which presents one row for each ground truth class and one column for each predicted class. Thus, the value of each cell of the such matrix (row $r$, column $c$) indicates the number of gestures belonging to the class $r$ that were classified as class $c$. 

Aiming at a fair comparison, for the GRIT dataset, the original evaluation procedure is executed, as described in \citep{tsironi2017analysis}. Thereby, we run five hold-out experiments. At each round, the dataset is shuffled and split into two subsets: a training subset (comprising 80\% of the dataset) and a testing subset (comprising 20\% of the dataset). Thus, rather than using a confusion matrix and the accuracy metrics to evaluate the model, we use a table with the metrics accuracy, recall, precision, and F1-score metrics for all classes, calculated as the average of each metric.

\section{Results and discussion}
\label{sec:discussion}

In this section, we present the results obtained for both datasets: Montalbano and GRIT.

\subsection{Results for the Montalbano Dataset}

\begin{figure}[!h]
\centering
\includegraphics [width = \linewidth]{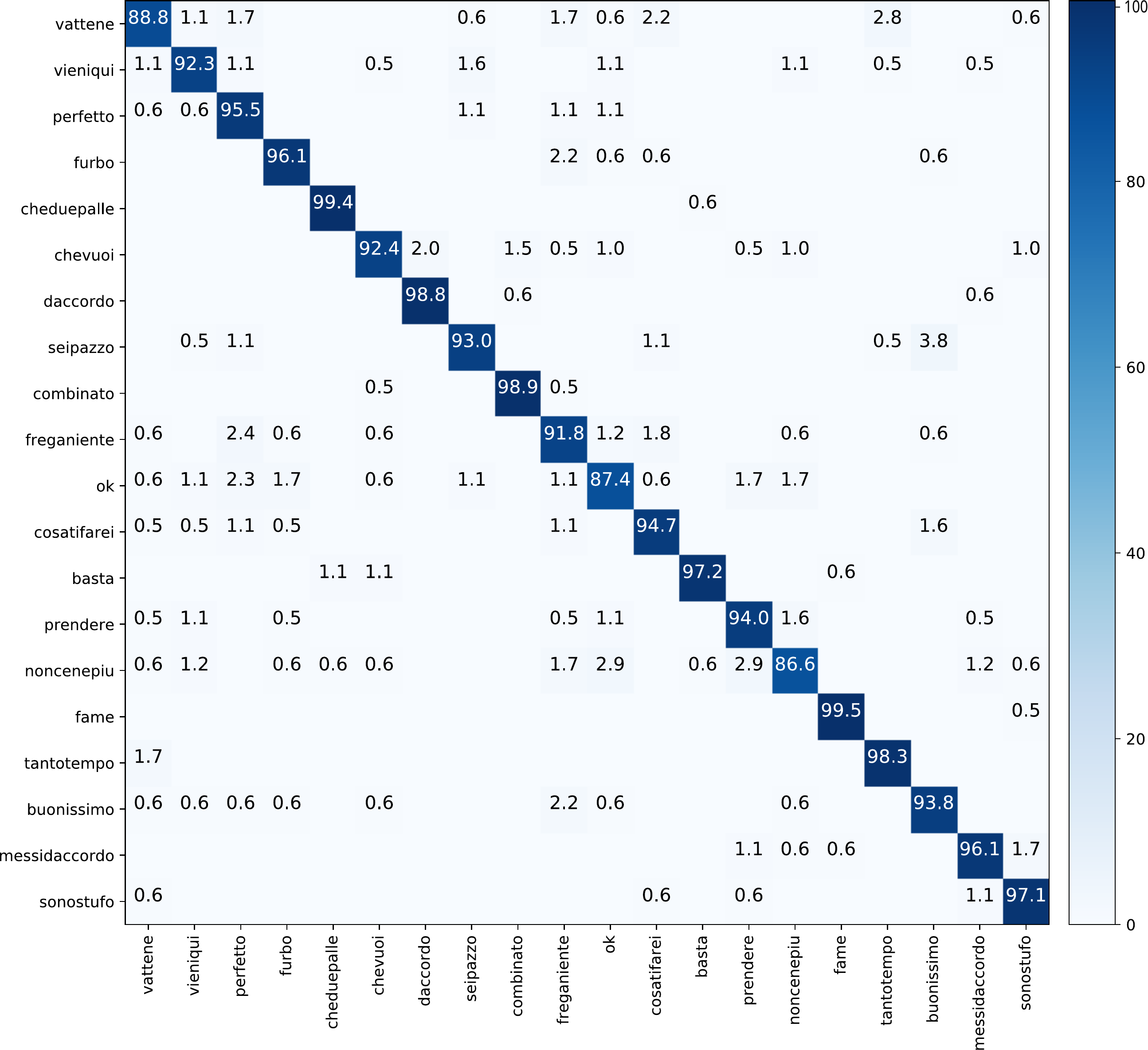}
\caption{Confusion matrix of the predictions made by the model trained with the Moltalbano gesture dataset.}
\label{fig:conf-matrix}
\end{figure}

The model trained according to the setup previously described achieves an average accuracy of 94.58\%, when executed over the testing dataset.  Table~\ref{tab:results} illustrates the accuracy obtained by each gesture and Figure \ref{fig:conf-matrix} shows the confusion matrix of the predictions. Note that for most gestures, the accuracy value is greater than 90\%, while for the gestures \textit{fame}, \textit{cheduepalle}, \textit{combinato}, \textit{daccordo} and \textit{tantotempo} the accuracy value is greater than 98\%. In just a few cases, like for the gestures \textit{vattene}, \textit{ok} and \textit{noncenepiu}, the accuracy is less than $90\%$. However, even for the worst case (\textit{noncenepiu}) the value of the accuracy is 89.63\% 

\begin{table}[!h] 
\centering
\caption{Result of the experiments on the Montalbano dataset.}
    \begin{tabular}{lclc}
        \hline 
        Gesture & Acc(\%) & Gesture & Acc(\%)\\
        \hline 
            fame & 99.46 & cosatifarei & 94.68 \\
            cheduepalle & 99.42 & prendere & 94.02 \\
            combinato & 98.91 & buonissimo & 93.82 \\
            daccordo & 98.77 & seipazzo & 92.97 \\
            tantotempo & 98.27 & chevuoi & 92.42 \\
            basta & 97.24 & vieniqui & 92.31 \\
            sonostufo & 97.14 & freganiente & 91.76 \\
            messidaccordo & 96.11 & vattene & 88.76 \\
            furbo & 96.07 & ok & 87.36 \\
            perfetto & 95.51 & noncenepiu & 86.63 \\ \hline
    \end{tabular} 
\label{tab:results}
 \end{table}

The original architecture is modified to demonstrate the impact of the soft-attention ensemble in the performance of the gesture classifier. In order to do that, we ran two experiments. Regarding the first one, we test different feature fusions: sum, mean and concatenation. The results of this experiment are summarized in Table \ref{tab:fusion}, where one can notice that the soft-attention ensemble achieves the best result, outperforming other techniques in more than $0.71\%$ of accuracy. 
For the second experiment, to show the influence of each CNN in each class, we remove the soft-attention ensemble, besides, the $4^{th}$ convolutional layer of each Resnet is directly linked to the fully connected layers. Hence, the gesture classifier using only Resnet $50$ achieved $91.95$\% of accuracy, meanwhile  Resnet $101$ achieved $90.70$\% of accuracy. Table~\ref{tab:results_separated} shows such results. Notice that, for the gestures \textit{fame}, \textit{cheduepalle}, \textit{combinato}, \textit{tantotempo}, and \textit{messidaccordo}, Resnet $50$ performed worse than Resnet $101$. However in other cases, for instance \textit{vattene}, \textit{freganiente} and \textit{cosatifarei}, Resnet $50$ outperforms Resnet $101$. Nevertheless for all gestures, the incorporations for the ensemble step results in the best accuracy, except for the gesture \textit{basta}, which Resnet 50 achieved $98.6\%$.

\begin{table}[!h] 
\centering
\caption{Accuracy for Montalbano dataset for different feature fusion techniques.}
\begin{tabular}{lclc}
\hline 
Feature Fusion & Accuracy\\
\hline 
Concatenation & 93.35\%  \\
Sum					 & 93.37\%   \\
Mean 				& 93.88\%  \\
\textbf{Soft-attention} & \textbf{94.58}\%  \\
\hline 
\end{tabular} 
\label{tab:fusion}
 \end{table}

\begin{table*}[!h] 
\centering
\caption{Result of the experiments on the Montalbano dataset using each Resnet individually and combining them through the soft-attention ensemble.}
    \begin{tabular}{lccc}
        \hline
        Gesture & Resnet 50 (\%) & Resnet 101 (\%) & Ensemble (\%)\\
        \hline 
            \textit{fame} & 96.76 & 97.84 & 99.46 \\
            \textit{cheduepalle} & 87.86 & 99.42 & 99.42 \\
            \textit{combinato} & 95.11 & 98.91 & 98.91 \\
            \textit{daccordo} & 98.16 & 98.77 & 98.77 \\
            \textit{tantotempo} & 96.53 & 98.27 & 98.27 \\
            \textit{basta} & 98.9 & 97.24 & 97.24 \\
            \textit{sonostufo} & 95.43 & 96.57 & 97.14 \\
            \textit{messidaccordo} & 92.22 & 96.11 & 96.11 \\
            \textit{furbo} & 94.94 & 91.57 & 96.07 \\
            \textit{perfetto} & 92.13 & 91.57 & 95.51 \\
            \textit{cosatifarei} & 92.02 & 81.91 & 94.68 \\
            \textit{prendere} & 91.85 & 92.93 & 94.02 \\
            \textit{buonissimo} & 91.57 & 91.01 & 93.82 \\
            \textit{seipazzo} & 92.97 & 90.81 & 92.97 \\
            \textit{chevuoi} & 91.41 & 91.92 & 92.42 \\
            \textit{vieniqui} & 87.91 & 84.07 & 92.31 \\
            \textit{freganiente} & 91.76 & 78.82 & 91.76 \\
            \textit{vattene} & 88.76 & 72.47 & 88.76 \\
            \textit{ok} & 83.33 & 82.76 & 87.36 \\
            \textit{noncenepiu} & 79.07 & 81.4 & 86.63 \\
         \hline
    \end{tabular} 
\label{tab:results_separated}
 \end{table*}
 
Thus, regarding the number of classes of the problem and the use of RGB information exclusively to represent and recognize dynamic gestures, we consider that the results presented here are competitive with other approaches that employ multimodal data. Furthermore, the method proposed in this paper entails one of the best results for classifying gestures on that dataset, outperforming other works that employ more than one source of information, for instance, \citep{Joshi2017}, \citep{CongqiCao2015} and  \citep{Escobedo-Cardenas2015}. 
A comparison among these different approaches, including our method, on the Montalbano dataset is shown in Table \ref{tab:montalbano_results}. As can be seen, our approach outperforms all the others.

 \begin{table}[!h]
 \caption{Comparing results between works that aimed to use the segmented videos of Montalbano dataset.}
 \label{tab:montalbano_results}
 \centering
\begin{tabular}{lllll}
\hline
\multicolumn{1}{c}{\textbf{Work}} & \multicolumn{1}{c}{\textbf{Used Data Type}} & \multicolumn{1}{c}{\textbf{Result}} \\
\hline
\citep{liu2019hidden} & Skeleton & 93.8\%\\
\citep{Efthimiou2016}   & Audio and RGB    &  93.00\%\\
\citep{Escobedo-Cardenas2015}   &  Skeleton and RGB-D &  88.38\%\\
\citep{Wu2016}    & Depth    & 82.62\%\\
\citep{Fernando2015}   & Skeleton and Audio &  80.29\%\\
\citep{CongqiCao2015}   & RGB  & 60.07\%\\
\textbf{our}   & \textbf{RGB}  & \textbf{94.58}\%\\

\hline
\end{tabular}
\end{table}

\subsection{Results for GRIT Dataset}

The results of the five hold-out experiments can be seen in Table~\ref{tab:grit_results}. As expected, our proposal achieves better results when considering the GRIT dataset, even outperforms the results in~\citep{tsironi2017analysis}. The improvement is more than $6\%$ for all metrics: accuracy, precision, recall, and F1-Score. Furthermore, our best result is $100\%$ of accuracy against $92.59\%$ achieved by the authors in~\citep{tsironi2017analysis}. This result implies that the star RGB indeed improves dynamic gesture representation and also can contribute to the dynamic gesture recognition research field. 

\begin{table}[!h]
\caption{Comparing results of our proposa against the results from \citep{tsironi2017analysis} using GRIT dataset.}
\label{tab:grit_results}
\centering
\begin{tabular}{lll}
\hline
Metric & Tsironi et al.~\citep{tsironi2017analysis} & Ours\\
\hline
Accuracy & $91.67\% \pm 1.13\%$ & $98.35\% \pm 0.90\%$\\
Precision & $92.35\% \pm 0.98\%$ & $98.55\% \pm 0.80\%$\\
Recall & $91.90\% \pm 1.05\%$ & $98.33\% \pm 0.95\%$\\
F1-score & $92.13\% \pm  1.00\%$ & $98.34\% \pm 0.93\%$\\
\hline
\end{tabular}
\end{table}

\subsection{General Comments}

Although the results are promising, some issues about our proposal should be considered. This technique must be used in an environment where the cameras are static or where the relative movement between the background and the person is minimal. Thus, when there is a moving camera, a possible solution could be to use homography estimation, through background features, for the purpose of calculating the motion of the camera, as proposed by~\citep{iDT}, for action recognition. The knowledge about the movements of the camera may allow to isolate the real movement of the person related to the gesture and then perform its recognition.

Also, as observed in \citep{tsironi2017analysis}, techniques like ours have the property of losing hand details. Thus, gestures that depend on the shape of the hands, and have the same movement of other gestures, could result in false-positives.  To illustrate the effect of this constraint, a technique of information visualization, also known as saliency map (Cam++~\citep{cam_pp})is applied to some gestures belonging to the class \textit{noncenepiu}, which are the ones referred to the worst results achieved by the classifier.

\begin{figure}[!h]
\centering
\subfloat[Resnet 50]{\includegraphics[width=0.23\linewidth]{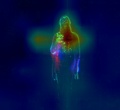}}\,
\subfloat[Resnet 101]{\includegraphics[width=0.23\linewidth]{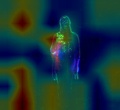}}\,
\subfloat[Ensemble - \textit{Ok} ($97.90\%$)]{\includegraphics[width=0.23\linewidth]{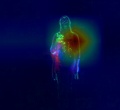}}\,
\subfloat[Ensemble - \textit{Noncenepiu} ($1.02\%$)]{\includegraphics[width=0.23 \linewidth]{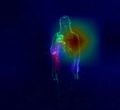}} \,

\subfloat[Resnet 50]{\includegraphics[width=0.23\linewidth]{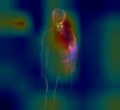}}\,
\subfloat[Resnet 101]{\includegraphics[width=0.23\linewidth]{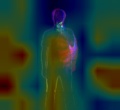}}\,
\subfloat[Ensemble - \textit{Freganiente} ($74.41\%$)]{\includegraphics[width=0.23\linewidth]{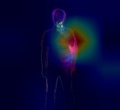}}\,
\subfloat[Ensemble - \textit{Noncenepiu} ($8.77\%$)]{\includegraphics[width=0.23\linewidth]{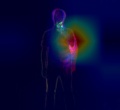}} \,

\subfloat[Resnet 50]{\includegraphics[width=0.23\linewidth]{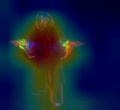}}\,
\subfloat[Resnet 101]{\includegraphics[width=0.23\linewidth]{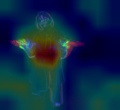}}\,
\subfloat[Ensemble - \textit{Basta} ($99.90\%$)]{\includegraphics[width=0.23\linewidth]{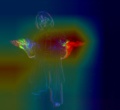}}\,
\subfloat[Ensemble - \textit{Noncenepiu} ($0.1\%$)]{\includegraphics[width=0.23 \linewidth]{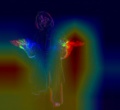}} \,

\subfloat[Resnet 50]{\includegraphics[width=0.23\linewidth]{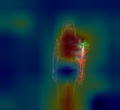}}\,
\subfloat[Resnet 101]{\includegraphics[width=0.23\linewidth]{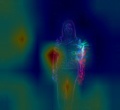}}\,
\subfloat[Ensemble - \textit{Prendere} ($87.38\%$)]{\includegraphics[width=0.23\linewidth]{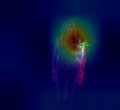}}\,
\subfloat[Ensemble - \textit{Noncenepiu} ($2.18\%$)]{\includegraphics[width=0.23 \linewidth]{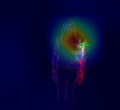}}\,
\caption{Salient map of the some gestures of the class \textit{noncenepiu} misclassified as another one.}
\label{fig:att_map}
\end{figure}

In general, the saliency maps of \textit{noncenepiu} (see Figure~\ref{fig:att_map}) are similar to the maps of the gestures \textit{ok}, \textit{freganiente}, \textit{basta} and \textit{prendere} (see the $15^{th}$ row of the confusion matrix in Figure~\ref{fig:conf-matrix}).

\begin{figure}[!h]
\centering
\subfloat[\textit{noncenepiu} - one hand]{\includegraphics[height =0.15\linewidth,width=0.15\linewidth]{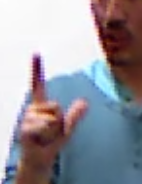}}\,
\subfloat[\textit{ok}]{\includegraphics[height =0.15\linewidth,width=0.15\linewidth]{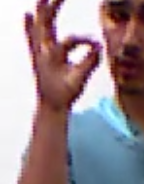}} \,
\subfloat[\textit{freganient}]{\includegraphics[height =0.15\linewidth,width=0.15\linewidth]{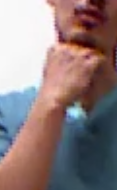}} \,
\subfloat[\textit{freganient}]{\includegraphics[height =0.15\linewidth,width=0.15\linewidth]{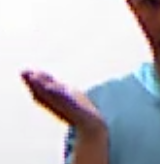}} \,

\subfloat[\textit{\textit{noncenepiu}} - two hands]{\includegraphics[width=0.3 \linewidth]{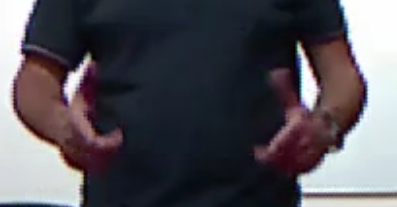}}\,
\subfloat[\textit{basta}]{\includegraphics[width=0.3\linewidth]{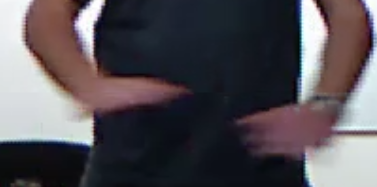}}\,

\caption{A sample of the hand shape of same gestures.}
\label{fig:hands_shape}
\end{figure}

From the videos, it is possible to see that the gesture \textit{noncenepiu} is done by twisting the arm over the elbow, with the thumb and pointer finger forming an \ASPADUPLA{L} (the hand-shape can be seen in Figures~\ref{fig:hands_shape}\textit{a} and~\ref{fig:hands_shape}\textit{e}). Therefore, when isolating this movement, for some individuals, it becomes very similar to the movement from the gestures \textit{ok}, \textit{prendere}, \textit{freganiente} and other ones. In this case, gesture recognition could be improved by using an approach capable of recognizing the hand details, instead of just the movements.

Besides, Figure~\ref{fig:att_map} shows how each CNN extracts different feature maps related to the input frame, whereas the soft-attention ensemble captures the essential information from each CNN. Other interesting observation regards the gesture \textit{basta}, which is confused with the gesture \textit{noncenepiu} when the user performs it with the two arms (see Figures~\ref{fig:att_map}-\textit{i}, \textit{j}, \textit{k}, \textit{l}). As discussed before, the movements are very similar even though the hand shapes are different from each other (see Figures~\ref{fig:hands_shape}~\textit{e}, \textit{f}).

\section{Conclusions and Future Work}
\label{sec:conclusions}


Considering the importance of dynamic gesture recognition for HMI and also the problem of recognizing gestures using just color information, 
this work reports two principal contributions: 
(\textit{i}) an approach called star RGB representation, that can describe and condense a video clip containing a dynamic gesture in only one RGB image; 
(\textit{ii}) a dynamic gesture classifier based in two pre-trained Resnets, a soft-attention ensemble followed by a set of fully connected layers. 
The experiments is carried out on both Montalbano and GRIT datasets, achieving an average accuracy of $94.58\%$ for the Montalbano dataset and a mean accuracy of $98.35\%$ over five randomly holdout experiments for the GRIT dataset.
The results obtained show that the star RGB, used with the soft-attention ensemble, outperforms previous works, such as ~\citep{tsironi2017analysis},~\citep{LiChuankun2017},~\citep{Joshi2017},~\citep{CongqiCao2015},~\citep{Wang2017}, achieving the state-of-the-art.

We suggest the following tasks as future works to improve the proposed approach. 
Firstly, the development of a new classifier architecture that takes into account the hand information of each gesture. 
Secondly, the application of this proposal in a robotic environment. For this, it is necessary to develop a spotting gesture algorithm that can operate in real-time. Then, using the proposed architecture, gestures of a new or different vocabulary could be recognized, allowing interaction with the robot.

\section*{Acknowledgments}

The authors would like to acknowledge the support from CAPES (Coordenação de Aperfeiçoamento de Pessoal de Nível Superior) through the scholarship given to the first author, as well as the support from NVIDIA Corporation through the donation of the Titan V GPU used in this research. 

\section*{Bibliography}
\bibliographystyle{elsarticle-num} 
\bibliography{referencies}

\begin{thebibliography}{10}
\expandafter\ifx\csname url\endcsname\relax
  \def\url#1{\texttt{#1}}\fi
\expandafter\ifx\csname urlprefix\endcsname\relax\def\urlprefix{URL }\fi
\expandafter\ifx\csname href\endcsname\relax
  \def\href#1#2{#2} \def\path#1{#1}\fi

\bibitem{mcneill1992hand}
D.~McNeill, Hand and mind: What gestures reveal about thought, University of
  Chicago press, 1992 (1992).

\bibitem{mitra2007survey}
S.~Mitra, T.~Acharya, Gesture recognition: A survey, IEEE Transactions on
  Systems, Man, and Cybernetics, Part C (Applications and Reviews) 37~(3)
  (2007) 311--324 (2007).
\newblock \href {https://doi.org/10.1109/TSMCC.2007.893280}
  {\path{doi:10.1109/TSMCC.2007.893280}}.

\bibitem{rautaray2015vision}
S.~S. Rautaray, A.~Agrawal, Vision based hand gesture recognition for human
  computer interaction: a survey, Artificial Intelligence Review 43~(1) (2015)
  1--54 (2015).
\newblock \href {https://doi.org/10.1007/s10462-012-9356-9}
  {\path{doi:10.1007/s10462-012-9356-9}}.

\bibitem{dynamic2016survey}
S.~Saikia, S.~Saharia, A survey on vision-based dynamic gesture recognition,
  International Journal of Computer Applications 138~(1) (2016).
\newblock \href {https://doi.org/10.5120/ijca2016908655}
  {\path{doi:10.5120/ijca2016908655}}.

\bibitem{neverova2016moddrop}
N.~Neverova, C.~Wolf, G.~Taylor, F.~Nebout, Moddrop: adaptive multi-modal
  gesture recognition, IEEE Transactions on Pattern Analysis and Machine
  Intelligence 38~(8) (2016) 1692--1706 (2016).
\newblock \href {https://doi.org/10.1109/TPAMI.2015.2461544}
  {\path{doi:10.1109/TPAMI.2015.2461544}}.

\bibitem{multimodal2017challenges}
S.~Escalera, V.~Athitsos, I.~Guyon, Challenges in multi-modal gesture
  recognition, in: Gesture Recognition, Springer, 2017, pp. 1--60 (2017).
\newblock \href {https://doi.org/10.1007/978-3-319-57021-1_1}
  {\path{doi:10.1007/978-3-319-57021-1_1}}.

\bibitem{barros2014real}
P.~Barros, G.~I. Parisi, D.~Jirak, S.~Wermter, Real-time gesture recognition
  using a humanoid robot with a deep neural architecture, in: Humanoid Robots
  (Humanoids), 2014 14th IEEE-RAS International Conference on, IEEE, 2014, pp.
  646--651 (2014).
\newblock \href {https://doi.org/10.1109/HUMANOIDS.2014.7041431}
  {\path{doi:10.1109/HUMANOIDS.2014.7041431}}.

\bibitem{liu2017survey}
W.~Liu, Z.~Wang, X.~Liu, N.~Zeng, Y.~Liu, F.~E. Alsaadi, A survey of deep
  neural network architectures and their applications, Neurocomputing 234
  (2017) 11--26 (2017).
\newblock \href {https://doi.org/10.1016/j.neucom.2016.12.038}
  {\path{doi:10.1016/j.neucom.2016.12.038}}.

\bibitem{3dConvolution_for_HAR}
S.~Ji, W.~Xu, M.~Yang, K.~Yu, 3d convolutional neural networks for human action
  recognition, IEEE transactions on pattern analysis and machine intelligence
  35~(1) (2013) 221--231 (2013).
\newblock \href {https://doi.org/10.1109/TPAMI.2012.59}
  {\path{doi:10.1109/TPAMI.2012.59}}.

\bibitem{3d_har2018}
K.~Hara, H.~Kataoka, Y.~Satoh, Can spatiotemporal 3d cnns retrace the history
  of 2d cnns and imagenet?, in: Proceedings of the IEEE conference on Computer
  Vision and Pattern Recognition, 2018, pp. 6546--6555 (2018).
\newblock \href {https://doi.org/10.1109/CVPR.2018.00685}
  {\path{doi:10.1109/CVPR.2018.00685}}.

\bibitem{zisserman2014}
K.~Simonyan, A.~Zisserman, Two-stream convolutional networks for action
  recognition in videos, in: Advances in neural information processing systems,
  2014, pp. 568--576 (2014).

\bibitem{zisserman2016}
C.~Feichtenhofer, A.~Pinz, A.~Zisserman, Convolutional two-stream network
  fusion for video action recognition, in: Proceedings of the IEEE conference
  on computer vision and pattern recognition, 2016, pp. 1933--1941 (2016).
\newblock \href {https://doi.org/10.1109/CVPR.2016.213}
  {\path{doi:10.1109/CVPR.2016.213}}.

\bibitem{zisserman2017}
J.~Carreira, A.~Zisserman, Quo vadis, action recognition? a new model and the
  kinetics dataset, in: proceedings of the IEEE Conference on Computer Vision
  and Pattern Recognition, 2017, pp. 6299--6308 (2017).
\newblock \href {https://doi.org/10.1109/CVPR.2017.502}
  {\path{doi:10.1109/CVPR.2017.502}}.

\bibitem{goodfellow2016deep}
I.~Goodfellow, Y.~Bengio, A.~Courville, Deep learning, MIT press, 2016 (2016).

\bibitem{escalera2013multi}
S.~Escalera, J.~Gonz{\`a}lez, X.~Bar{\'o}, M.~Reyes, O.~Lopes, I.~Guyon,
  V.~Athitsos, H.~Escalante, Multi-modal gesture recognition challenge 2013:
  Dataset and results, in: Proceedings of the 15th ACM on International
  conference on multimodal interaction, ACM, 2013, pp. 445--452 (2013).
\newblock \href {https://doi.org/10.1145/2522848.2532595}
  {\path{doi:10.1145/2522848.2532595}}.

\bibitem{escalera2014}
S.~Escalera, X.~Bar{\'o}, J.~Gonzalez, M.~A. Bautista, M.~Madadi, M.~Reyes,
  V.~Ponce-L{\'o}pez, H.~J. Escalante, J.~Shotton, I.~Guyon, Chalearn looking
  at people challenge 2014: Dataset and results, in: Workshop at the European
  Conference on Computer Vision, Springer, 2014, pp. 459--473 (2014).
\newblock \href {https://doi.org/10.1007/978-3-319-16178-5_32}
  {\path{doi:10.1007/978-3-319-16178-5_32}}.

\bibitem{Efthimiou2016}
E.~Efthimiou, S.~E. Fotinea, T.~Goulas, M.~Koutsombogera, P.~Karioris,
  A.~Vacalopoulou, I.~Rodomagoulakis, P.~Maragos, C.~Tzafestas, V.~Pitsikalis,
  Y.~Koumpouros, A.~Karavasili, P.~Siavelis, F.~Koureta, D.~Alexopoulou, {The
  MOBOT rollator human-robot interaction model and user evaluation process},
  2016 IEEE Symposium Series on Computational Intelligence (SSCI) (2016) 1--8
  (2016).
\newblock \href {https://doi.org/10.1109/SSCI.2016.7850061}
  {\path{doi:10.1109/SSCI.2016.7850061}}.

\bibitem{LiChuankun2017}
C.~Li, Y.~Hou, P.~Wang, W.~Li, Joint distance maps based action recognition
  with convolutional neural networks, IEEE Signal Processing Letters 24~(5)
  (2017) 624--628 (2017).
\newblock \href {https://doi.org/10.1109/LSP.2017.2678539}
  {\path{doi:10.1109/LSP.2017.2678539}}.

\bibitem{Joshi2017}
A.~Joshi, C.~Monnier, M.~Betke, S.~Sclaroff, {Comparing random forest
  approaches to segmenting and classifying gestures}, Image and Vision
  Computing 58 (2017) 86--95 (2017).
\newblock \href {https://doi.org/10.1016/j.imavis.2016.06.001}
  {\path{doi:10.1016/j.imavis.2016.06.001}}.

\bibitem{Wang2017}
H.~Wang, L.~Wang, Modeling temporal dynamics and spatial configurations of
  actions using two-stream recurrent neural networks, in: Proceedings of the
  IEEE Conference on Computer Vision and Pattern Recognition, 2017, pp.
  499--508 (2017).
\newblock \href {https://doi.org/10.1109/CVPR.2017.387}
  {\path{doi:10.1109/CVPR.2017.387}}.

\bibitem{Neverova2015}
N.~Neverova, C.~Wolf, G.~W. Taylor, F.~Nebout, Multi-scale deep learning for
  gesture detection and localization, in: Workshop at the European conference
  on computer vision, Springer, 2014, pp. 474--490 (2014).
\newblock \href {https://doi.org/10.1007/978-3-319-16178-5_33}
  {\path{doi:10.1007/978-3-319-16178-5_33}}.

\bibitem{Georgios2014}
G.~Pavlakos, S.~Theodorakis, V.~Pitsikalis, A.~Katsamanis, P.~Maragos,
  Kinect-based multimodal gesture recognition using a two-pass fusion scheme,
  in: 2014 IEEE International Conference on Image Processing (ICIP), IEEE,
  2014, pp. 1495--1499 (2014).
\newblock \href {https://doi.org/10.1109/ICIP.2014.7025299}
  {\path{doi:10.1109/ICIP.2014.7025299}}.

\bibitem{Pigou2018}
L.~Pigou, A.~Van Den~Oord, S.~Dieleman, M.~Van~Herreweghe, J.~Dambre, Beyond
  temporal pooling: Recurrence and temporal convolutions for gesture
  recognition in video, International Journal of Computer Vision 126~(2-4)
  (2018) 430--439 (2018).
\newblock \href {https://doi.org/10.1007/s11263-016-0957-7}
  {\path{doi:10.1007/s11263-016-0957-7}}.

\bibitem{lstm}
S.~Hochreiter, J.~Schmidhuber, Long short-term memory, Neural computation 9~(8)
  (1997) 1735--1780 (1997).
\newblock \href {https://doi.org/10.1162/neco.1997.9.8.1735}
  {\path{doi:10.1162/neco.1997.9.8.1735}}.

\bibitem{liu20193d}
X.~Liu, G.~Zhao, 3d skeletal gesture recognition via sparse coding of
  time-warping invariant riemannian trajectories, in: International Conference
  on Multimedia Modeling, Springer, 2019, pp. 678--690 (2019).
\newblock \href {https://doi.org/10.29007/xhfp} {\path{doi:10.29007/xhfp}}.

\bibitem{liu2019hidden}
X.~Liu, H.~Shi, X.~Hong, H.~Chen, D.~Tao, G.~Zhao, Hidden states exploration
  for 3d skeleton-based gesture recognition, in: 2019 IEEE Winter Conference on
  Applications of Computer Vision (WACV), IEEE, 2019, pp. 1846--1855 (2019).
\newblock \href {https://doi.org/10.1109/WACV.2019.00201}
  {\path{doi:10.1109/WACV.2019.00201}}.

\bibitem{ChenXi2014}
X.~Chen, M.~Koskela, {Using appearance-based hand features for dynamic RGB-D
  gesture recognition}, Proceedings - International Conference on Pattern
  Recognition (2014) 411--416 (2014).
\newblock \href {https://doi.org/10.1109/ICPR.2014.79}
  {\path{doi:10.1109/ICPR.2014.79}}.

\bibitem{Yao2014}
A.~Yao, L.~V. Gool, P.~Kohli, {Gesture recognition portfolios for
  personalization}, Proceedings of the IEEE Computer Society Conference on
  Computer Vision and Pattern Recognition (2014) 1923--1930 (2014).
\newblock \href {https://doi.org/10.1109/CVPR.2014.247}
  {\path{doi:10.1109/CVPR.2014.247}}.

\bibitem{Wu2014}
D.~Wu, L.~Shao, {Leveraging hierarchical parametric networks for skeletal
  joints based action segmentation and recognition}, Proc. IEEE Comput. Soc.
  Conf. Comput. Vis. Pattern Recognit. (2014) 724--731 (2014).
\newblock \href {https://doi.org/10.1109/CVPR.2014.98}
  {\path{doi:10.1109/CVPR.2014.98}}.

\bibitem{Fernando2015}
B.~Fernando, E.~Gavves, M.~{Jos{\'{e}} Oramas}, A.~Ghodrati, T.~Tuytelaars,
  {Modeling video evolution for action recognition}, Proceedings of the IEEE
  Computer Society Conference on Computer Vision and Pattern Recognition
  07-12-June (2015) 5378--5387 (2015).
\newblock \href {https://doi.org/10.1109/CVPR.2015.7299176}
  {\path{doi:10.1109/CVPR.2015.7299176}}.

\bibitem{Escobedo-Cardenas2015}
E.~Escobedo-Cardenas, G.~Camara-Chavez, {A robust gesture recognition using
  hand local data and skeleton trajectory}, in: 2015 IEEE International
  Conference on Image Processing (ICIP), IEEE, 2015, pp. 1240--1244 (sep 2015).
\newblock \href {https://doi.org/10.1109/ICIP.2015.7350998}
  {\path{doi:10.1109/ICIP.2015.7350998}}.

\bibitem{Wu2016}
D.~Wu, L.~Pigou, P.~J. Kindermans, N.~D.~H. Le, L.~Shao, J.~Dambre, J.~M.
  Odobez, {Deep Dynamic Neural Networks for Multimodal Gesture Segmentation and
  Recognition}, IEEE Transactions on Pattern Analysis and Machine Intelligence
  38~(8) (2016) 1583--1597 (2016).
\newblock \href {https://doi.org/10.1109/TPAMI.2016.2537340}
  {\path{doi:10.1109/TPAMI.2016.2537340}}.

\bibitem{CongqiCao2015}
C.~Cao, Y.~Zhang, H.~Lu, Multi-modal learning for gesture recognition, in: 2015
  IEEE International Conference on Multimedia and Expo (ICME), IEEE, 2015, pp.
  1--6 (2015).
\newblock \href {https://doi.org/10.1109/ICME.2015.7177460}
  {\path{doi:10.1109/ICME.2015.7177460}}.

\bibitem{MHIrepresentation}
A.~F. Bobick, J.~W. Davis, The recognition of human movement using temporal
  templates, IEEE Transactions on Pattern Analysis \& Machine Intelligence~(3)
  (2001) 257--267 (2001).
\newblock \href {https://doi.org/10.1109/34.910878}
  {\path{doi:10.1109/34.910878}}.

\bibitem{tsironi2017analysis}
E.~Tsironi, P.~Barros, C.~Weber, S.~Wermter, An analysis of convolutional long
  short-term memory recurrent neural networks for gesture recognition,
  Neurocomputing 268 (2017) 76--86 (2017).
\newblock \href {https://doi.org/10.1016/j.neucom.2016.12.088}
  {\path{doi:10.1016/j.neucom.2016.12.088}}.

\bibitem{samatelo2012new}
J.~L.~A. Samatelo, E.~O.~T. Salles, A new change detection algorithm for visual
  surveillance system, IEEE Latin America Transactions 10~(1) (2012) 1221--1226
  (2012).
\newblock \href {https://doi.org/10.1109/TLA.2012.6142465}
  {\path{doi:10.1109/TLA.2012.6142465}}.

\bibitem{resnet}
K.~He, X.~Zhang, S.~Ren, J.~Sun, Deep residual learning for image recognition,
  in: Proceedings of the IEEE conference on computer vision and pattern
  recognition, 2016, pp. 770--778 (2016).
\newblock \href {https://doi.org/10.1109/CVPR.2016.90}
  {\path{doi:10.1109/CVPR.2016.90}}.

\bibitem{russakovsky2015imagenet}
O.~Russakovsky, J.~Deng, H.~Su, J.~Krause, S.~Satheesh, S.~Ma, Z.~Huang,
  A.~Karpathy, A.~Khosla, M.~Bernstein, et~al., Imagenet large scale visual
  recognition challenge, International Journal of Computer Vision 115~(3)
  (2015) 211--252 (2015).
\newblock \href {https://doi.org/10.1007/s11263-015-0816-y}
  {\path{doi:10.1007/s11263-015-0816-y}}.

\bibitem{simonyan2014vgg}
K.~Simonyan, A.~Zisserman, Very deep convolutional networks for large-scale
  image recognition, arXiv preprint arXiv:1409.1556 (2014).

\bibitem{guo2016densenet}
Y.~Guo, Y.~Liu, A.~Oerlemans, S.~Lao, S.~Wu, M.~S. Lew, Deep learning for
  visual understanding: A review, Neurocomputing 187 (2016) 27--48 (2016).
\newblock \href {https://doi.org/10.1016/j.neucom.2015.09.116}
  {\path{doi:10.1016/j.neucom.2015.09.116}}.

\bibitem{relu2010rectified}
V.~Nair, G.~E. Hinton, Rectified linear units improve restricted boltzmann
  machines, in: Proceedings of the 27th international conference on machine
  learning (ICML-10), 2010, pp. 807--814 (2010).

\bibitem{ioffe2015batch}
S.~Ioffe, C.~Szegedy, Batch normalization: Accelerating deep network training
  by reducing internal covariate shift, arXiv preprint arXiv:1502.03167 (2015).

\bibitem{srivastava2014dropout}
N.~Srivastava, G.~Hinton, A.~Krizhevsky, I.~Sutskever, R.~Salakhutdinov,
  Dropout: A simple way to prevent neural networks from overfitting, The
  Journal of Machine Learning Research 15~(1) (2014) 1929--1958 (2014).

\bibitem{pytorch}
A.~Paszke, S.~Gross, S.~Chintala, G.~Chanan, E.~Yang, Z.~DeVito, Z.~Lin,
  A.~Desmaison, L.~Antiga, A.~Lerer, Automatic differentiation in pytorch
  (2017).

\bibitem{iDT}
H.~Wang, C.~Schmid, Action recognition with improved trajectories, in:
  Proceedings of the IEEE international conference on computer vision, 2013,
  pp. 3551--3558 (2013).
\newblock \href {https://doi.org/10.1109/ICCV.2013.441}
  {\path{doi:10.1109/ICCV.2013.441}}.

\bibitem{cam_pp}
A.~Chattopadhay, A.~Sarkar, P.~Howlader, V.~N. Balasubramanian, Grad-cam++:
  Generalized gradient-based visual explanations for deep convolutional
  networks, in: Applications of Computer Vision (WACV), 2018 IEEE Winter
  Conference on, IEEE, 2018, pp. 839--847 (2018).
\newblock \href {https://doi.org/10.1109/WACV.2018.00097}
  {\path{doi:10.1109/WACV.2018.00097}}.

\end{thebibliography}

 \authorbiography[height={2.3cm},width={2.5cm}, overhang=0pt, imagewidth=3.5cm,wraplines=8,imagepos=l]{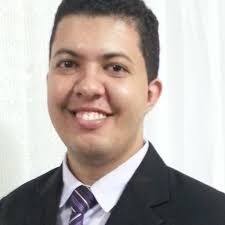}{}{\textbf{Clebeson Canuto dos Santos} is a Ph.D. student in Electrical Engineering at Federal University of Espirito Santo, Brazil, advised by Dr. Raquel Vassallo. He received his M.S. in Computer Science in  2016  from  Federal University of Sergipe, Brazil,  advised by Dr. Eduardo Freire. He earned his B.S. in Information System in 2013 from University Tiradentes, Brazil. His research interests are mainly in computer vision, deep learning and human-robot interaction.

}
\authorbiography[height={2.3cm},width={2.2cm}, overhang=0pt, imagewidth=3.5cm,wraplines=10,imagepos=l]{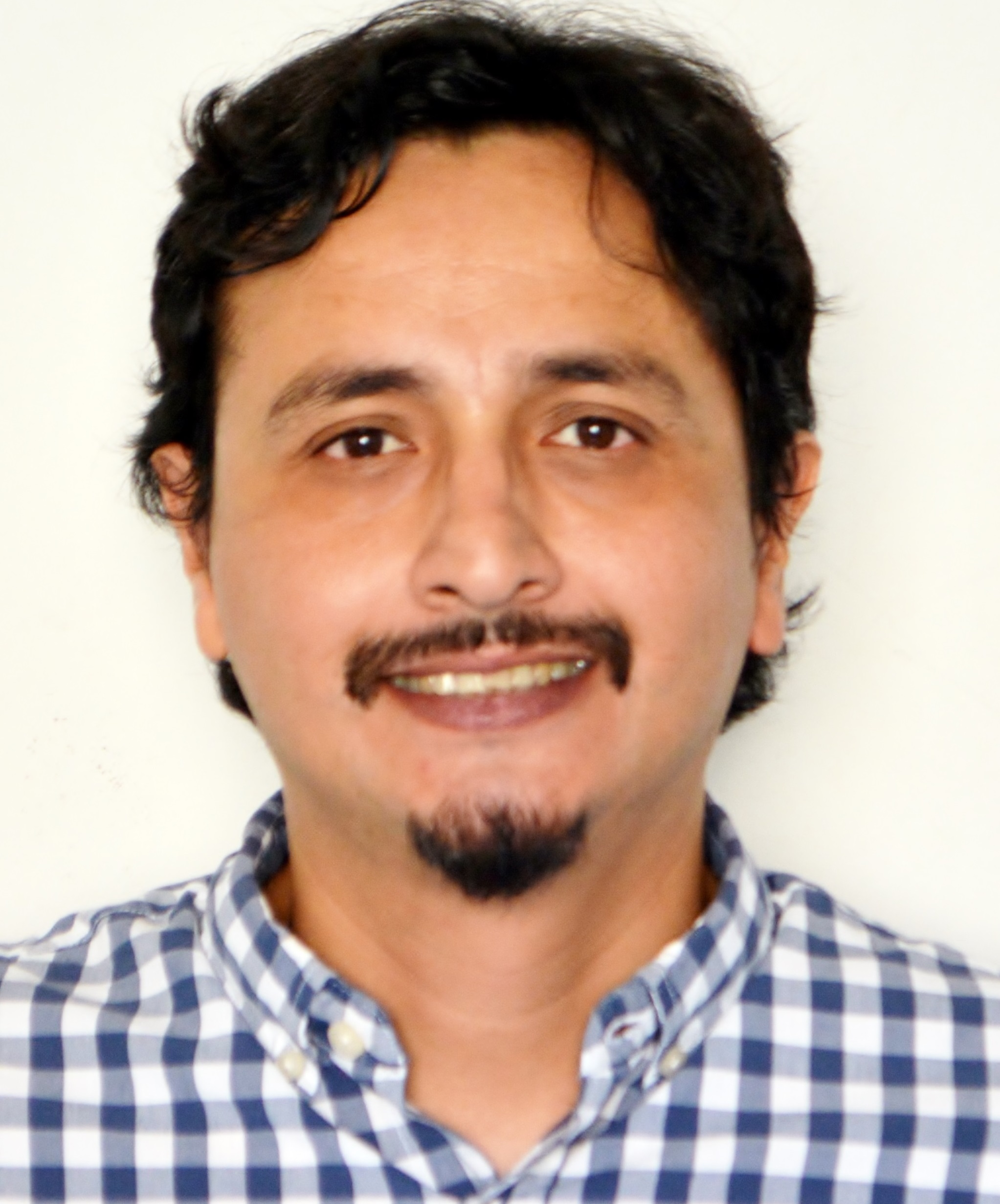}{}{\textbf{Jorge Leonid Aching Samatelo} has a degree in Electronic Engineering from the National University of San Marcos (2001), Peru,a M.S. in Electrical Engineering from Federal University of Esp\'{i}rito Santo (2007), Brazil, and a Ph.D. in Electrical Engineering from Federal University of Esp\'{i}rito Santo (2012), Brazil.  He is currently Associate Professor at the Federal University of Esp\'{i}rito Santo. 
His areas of expertise are image processing, computer vision, and deep learning.
}
\authorbiography[height={2.7cm},width={2.2cm}, overhang=0pt, imagewidth=3.5cm,wraplines=10,imagepos=l]{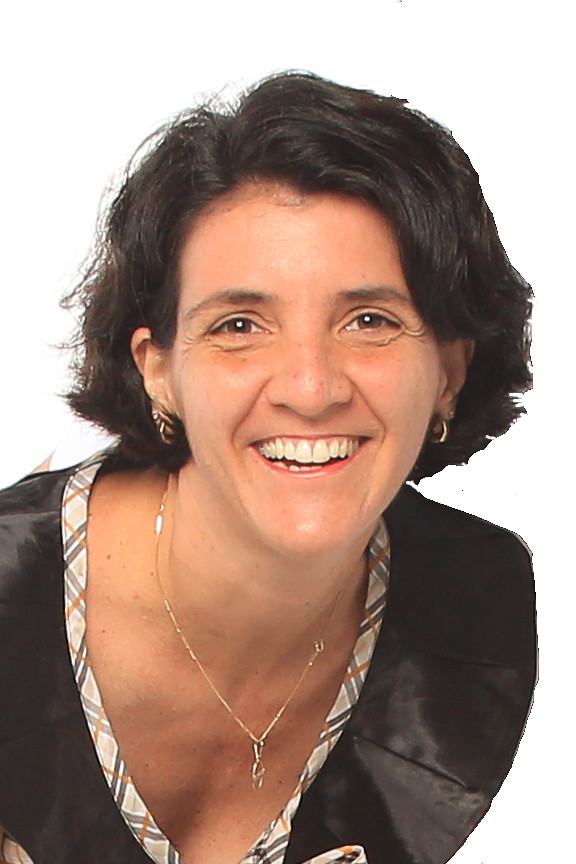}{}{\textbf{Raquel Frizera Vassallo} received the M.Sc. degree in automation and the Ph.D. degree in electrical engineering from the Federal University of Espirito Santo (UFES), Brazil, in 1998 and 2004, respectively. Her Ph.D. dissertation was on computer vision applied to mobile robotics in cooperation with the Instituto Superior Técnico, Lisbon, Portugal. She is currently an Associate Professor with the Electrical Engineering Department, UFES. Her research is mainly focused on computer vision for mapping, robot navigation, intelligent spaces, and mobile robotics.}

 \Authorbiography[totoc=true,BiographyName={About the Authors}]

\end{document}